\documentclass[lettersize,journal]{IEEEtran}

\IEEEoverridecommandlockouts                        

\usepackage[accsupp]{axessibility}
\usepackage{graphicx}
\usepackage{graphics} 
\usepackage{booktabs}
\usepackage{comment}
\usepackage{float}
\usepackage{amsmath} 
\usepackage{amssymb} 
\usepackage{multirow}
\usepackage[dvipsnames]{xcolor}
\usepackage{algorithm,algorithmic} 
\usepackage{color, colortbl}
\usepackage{cellspace}
\usepackage{url}
\usepackage{adjustbox}
\usepackage{subfigure}
\usepackage{tabularx}
\usepackage{float}
\usepackage{cite}
\usepackage{hyperref}
\newcommand{\keywordsnew}[1]{\textbf{\textit{keywords: }}{#1}}

\renewcommand{\theenumi}{\Alph{enumi}}

\definecolor{LightCyan}{rgb}{0.88,1,1}
\definecolor{gray}{RGB}{240, 240, 240}
\definecolor{dgray}{RGB}{105, 105, 105}

\usepackage{caption}
\title{\LARGE \bf Efficient Baselines for Motion Prediction in Autonomous Driving}

\author{Carlos Gómez-Huélamo~$^{1}$, Marcos V. Conde~$^{2}$, Rafael Barea~$^{1}$, Manuel Ocaña~$^{1}$ and Luis M. Bergasa~$^{1}$
    \thanks{$^{1}$~Carlos Gómez-Huélamo, Rafael Barea, Manuel Ocaña and Luis M. Bergasa ((carlos.gomezh, rafael.barea, manuel.ocanna, luism.bergasa)@uah.es) are with the Department of Electronics, University of Alcal{\'a} (UAH), Spain. \\
    $^{2}$~Marcos Conde (marcos.conde@uni-wuerzburg.de) is with the University of Würzburg, Computer Vision Lab, Germany.
    }
}

\begin{document}
\bstctlcite{IEEEexample:BSTcontrol}

\maketitle
\thispagestyle{empty}
\pagestyle{empty}


\begin{abstract}
Motion Prediction (MP) of multiple surroundings agents is a crucial task in arbitrarily complex environments, from simple robots to Autonomous Driving Stacks (ADS). Current techniques tackle this problem using end-to-end pipelines, where the input data is usually a rendered top-view of the physical information and the past trajectories of the most relevant agents; leveraging this information is a must to obtain optimal performance. In that sense, a reliable ADS must produce reasonable predictions on time. However, despite many approaches use simple ConvNets and LSTMs to obtain the social latent features, State-Of-The-Art (SOTA) models might be too complex for real-time applications when using both sources of information (map and past trajectories) as well as little interpretable, specially considering the physical information. Moreover, the performance of such models highly depends on the number of available inputs for each particular traffic scenario, which are expensive to obtain, particularly, annotated High-Definition (HD) maps. 

In this work, we propose several efficient baselines for the well-known Argoverse 1 Motion Forecasting Benchmark. We aim to develop compact models using SOTA techniques for MP, including attention mechanisms and GNNs. Our lightweight models use standard social information and interpretable map information such as points from the driveable area and plausible centerlines by means of a novel preprocessing step based on kinematic constraints, in opposition to black-box CNN-based or too-complex graphs methods for map encoding, to generate plausible multimodal trajectories achieving up-to-pair accuracy with less operations and parameters than other SOTA methods. Our code is publicly available at \href{https://github.com/Cram3r95/mapfe4mp}{https://github.com/Cram3r95/mapfe4mp}.
\end{abstract}

\keywordsnew{Autonomous Driving, Argoverse, Trajectory Forecasting, Motion Prediction, Multi-agent, Deep Learning}

\section{Introduction}
\label{sec:introduction}

Autonomous Driving (AD) is one of the most challenging research topics in academia and industry due to its real-world impact: improved safety, reduced congestion and greater mobility~\cite{KATRAKAZAS2015416realtime, vitelli2021safetynet}. Predicting the future behavior of traffic agents around the ego-vehicle is one of the key unsolved challenges in reaching full self-driving autonomy~\cite{kesten2019lyft, houston2020onelyftdata}. In that sense, an \textbf{Autonomous Driving Stack (ADS)} can be hierarchically broken down in the following tasks: (i) perception, responsible for identifying what is around us, then track and \textbf{predict} what will happen next, (ii) planning and decision-making, deciding what the AD stack is going to do in the near future and (iii) control, that sends the corresponding low-level commands (brake, throttle and steering angle) to the vehicle. 

\begin{figure}[!ht]
  \centering
   \includegraphics[trim={0 2cm 0 0}, clip, width=\linewidth]{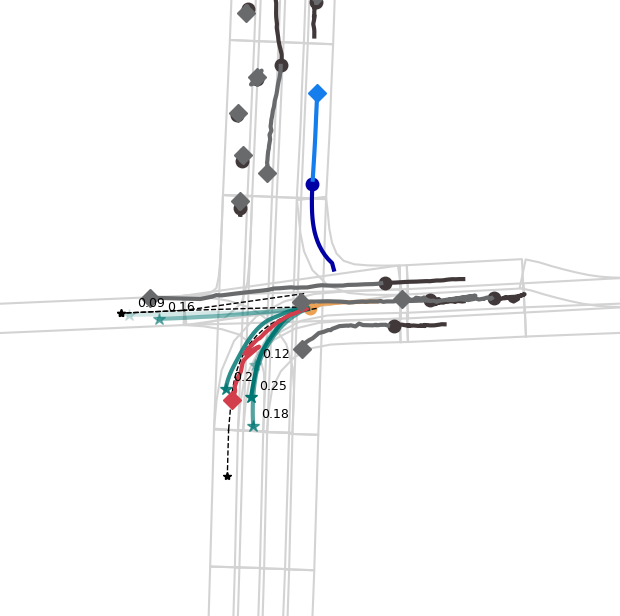}
   \caption{Motion Prediction Scenario in \textbf{Argoverse 1}~\cite{chang2019argoverse}. We represent: our vehicle (\textbf{\textcolor{blue}{ego}}), the \textbf{\textcolor{YellowOrange}{target agent}}, and \textbf{\textcolor{dgray}{other agents}}. We can also see the \textbf{\textcolor{red}{ground-truth}} trajectory of the target agent, our \textbf{\textcolor{ForestGreen}{multimodal predictions}} (with the corresponding confidences) and \textbf{plausible centerlines}. Circles represent last observations and diamonds last future positions.}
   \label{fig:results_teaser}
\end{figure}

Assuming the surrounding agents have been detected and tracked (we have their past trajectories), the final task of the perception layer is known as Motion Prediction (MP), that is, predicting the future trajectories ~\cite{mahjourian2022occupancy, salzmann2020trajectron++} of the surrounding traffic agents given the past on-board sensor and map information, and taking into account the corresponding traffic rules and social interaction among the agents. 

These predictions are required to be \textbf{multi-modal}, which means given the past motion of a particular vehicle and its surrounding scene, there may exist more than one possible future behaviour (also known as modes). Therefore, MP models need to cover the different choices a driver could make (i.e. going straight or turning, accelerations or slowing down) as a possible trajectory in the immediate future or as a probability distribution of the agent's future location~\cite{gilles2021home, dendorfer2020goalgan}. In other words, when an AD stack attempts to make a specific action (e.g. left turn), it must consider the future motion of the other vehicles, since the own future motion (also known as motion planning) depends on the all possible maneuvers of the other agents of the scene for safe driving. Fig. \ref{fig:results_teaser} shows this concept.

Traditional methods for motion forecasting~\cite{KATRAKAZAS2015416realtime, gomez2021smartmot} are based on physical kinematic constraints and road map information with handcrafted rules. Though these approaches are sufficient in many simple situations (i.e. cars moving in constant velocity), they fail to capture the rich behavior strategies and interaction in complex scenarios, in such a way they are only suitable for simple prediction scenes and short-time prediction tasks \cite{huang2022survey}.\\

On the other hand, recently Deep Learning (DL) based methods have dominated this task and they usually follow an encoder-decoder paradigm. The advances in DL allow us to understand and capture the complexity of a driving scenario using data-driven methods~\cite{deo2018convolutionalmotion, casas2018intentnet, chai2019multipath} and achieve the most promising \textit{state-of-the-art} results by learning such intrinsic rules. Methods based on Graph Neural Networks (GNNs) \cite{gilles2021gohome} \cite{liang2020learninggraph} have achieved SOTA results on the most relevant benchmarks for Motion Prediction, though these methods lack interpretability and control regarding the graph-based modules \cite{huang2022multi}. Moreover, despite  Generative Adversarial Network (GAN) approaches~\cite{sadeghian2019sophie, dendorfer2020goalgan, gupta2018sgan, gomez2022exploring} provide certain control since they are focused on more simple methods framed in an adversarial training, most competitive approaches on self-driving MP benchmarks such as Argoverse \cite{chang2019argoverse}, NuScenes \cite{caesar2020nuscenes} or Waymo \cite{Sun_2020_CVPR_waymodata} \textbf{do not} use adversarial training.

In these models, an encoder usually takes into account multiple-agents history states (position, velocity, etc.) as well as the local High Definition (HD) Map~\cite{can2022maps}, where obtaining and fusing this information (actor-to-actor, map-to-actor, map-to-map, actor-to-map, etc.) is a research topic by itself~\cite{varadarajan2021multipath++, zeng2021lanercnn, liang2020learninggraph} and a core part in the AD pipeline. In that sense, researchers have identified a bottleneck for efficient real-time applications~\cite{KATRAKAZAS2015416realtime, gomez2021smartmot}, as usually, more data-inputs implies higher complexity and inference time~\cite{gao2020vectornet}.
As observed, DL methods are data-hungry and use complex features (both social and physical) to predict, in an accurate way, the future behaviour of the agents in the scene. Most \textit{state-of-the-art} MP methods require an overwhelmed amount of information as input, specially in terms of the physical context -aforementioned HD maps-. This might be inefficient in terms of latency, parallelism or computation ~\cite{gao2020vectornet, walters2020trajectory}. \\

In this paper, following the same principles as recent SOTA methods, we aim to achieve competitive results that ensure reliable predictions, as observed in Fig.~\ref{fig:results_teaser}, yet, using \textbf{light-weight} attention-based models that take as input the past trajectories of each agent, and integrate prior-knowledge about the map easily. The main contributions of our work are as following: 

\begin{itemize}
    \item (1) Identify a key problem in the size of motion prediction models, with implications in real-time inference and edge-device deployment.
    \item (2) Propose several efficient baselines for vehicle motion prediction that do not explicitly rely on an exhaustive analysis of the context HD map (either vectorized or rasterized), but on prior map information obtained in a simple preprocessing step, that serves as a guide in the prediction.
    \item (3) Use fewer parameters and operations (FLOPs) than other SOTA models to achieve competitive performance on Argoverse 1.0~\cite{chang2019argoverse} with lower computational cost.
\end{itemize} 

\section{Related work}
\label{sec:related_work}

One of the crucial tasks that Autonomous Vehicles (AV) must face during navigation, specially in arbitrarily complex urban scenarios, is to predict the behaviour of dynamic obstacles \cite{chang2019argoverse, salzmann2020trajectron++}. In a similar way to humans that pay more attention to close obstacles and upcoming turns, rather than considering the obstacles far away, the perception layer of an AD stack must focus more on the salient regions of the scene, and the more relevant agents to predict the future behaviour of each traffic participant. 

Most traditional predictions methods \cite{huang2022survey}, which usually only consider physics-related factors (like the velocity and acceleration of the target vehicle that is going to be predicted) and road-related factors (prediction as close as possible to the road centerline), are only suitable for short-time prediction tasks \cite{huang2022survey} and simple traffic scenarios, such as constant velocity (CV) in a highway or a curve (Constant Turn Rate Velocity, CTRV) where a single path is allowed, i.e. multiple choices computation are not required. Recently, MP methods based on DL have become increasingly popular since they are able not only to take into account these above-mentioned factors but also consider \textbf{interaction-related factors} (like agent-agent \cite{gupta2018social}, agent-map \cite{casas2018intentnet} and map-map \cite{liang2020learninggraph}) in such a way the algorithm can adapt to more complex traffic scenarios (intersections, sudden breaks and accelerations, etc.). It must be consider that multimodal, specially in the field of vehicle motion prediction, does not refer necessarily to different directions (e.g. turn to the left, turn to the right, continue forward in an intersection), but it may refer to different predictions in the same direction that model a sudden positive or negative acceleration, so as to imitate a realistic human behaviour in complex situations. As expected, neither classical nor machine learning (ML) methods can model these situations \cite{huang2022survey}.

\begin{table*}[t]
\begin{center}
\caption{\textbf{State-of-the-art methods for Motion Prediction}. Main categories are Encoder (splitted into motion history, social info (agent interactions) and map info (physical information)), Decoder, Output representation and Distribution over future trajectories.}
\small
\begin{tabular}{r |ccc|c|c|c}
\toprule
\textbf{Method}	&	& \textbf{Encoder}	&	& \textbf{Decoder}	& \textbf{Output}  & \textbf{Trajectory Distribution}	\\
	& Motion history	& Social info	& Map info	&	&	&	\\
\midrule
\midrule
SocialLSTM~\cite{alahi2016social}	& LSTM	& spatial pooling	& --	& LSTM	& states	& samples	\\
SocialGan~\cite{gupta2018social}	& LSTM	& maxpool	& --	& LSTM	& states	& samples	\\
Jean~\cite{mercat2020multiattentmotion}	& LSTM	& attention	& --	& LSTM	& states	& GMM	\\
TNT~\cite{zhao2020tnt}	& polyline	& maxpool, attention	& polyline	& MLP	& states	& weighted set	\\
LaneGCN~\cite{liang2020learninggraph}	& 1D-conv	& GNN	& GNN	& MLP	& states	& weighted set	\\
WIMP~\cite{khandelwal2020if}	& LSTM	& GNN+attention	& polyline	& LSTM	& states	& GMM	\\
VectorNet~\cite{gao2020vectornet}	& polyline	& maxpool, attention	& polyline	& MLP	& states	& unimodal	\\
SceneTransformer~\cite{ngiam2021scene}	& attention	& attention	& polyline	& attention	& states	& weighted set	\\
HOME~\cite{gilles2021home}	& raster	& attention	& raster	& conv	& states	& heatmap	\\
GOHOME~\cite{gilles2021gohome}	& 1D-conv+GRU	& GNN	& GNN	& MLP	& states	& heatmap	\\
MP3~\cite{casas2021mp3}	& raster	& conv	& raster	& conv	& cost function	& weighted samples	\\
ExploringGAN~\cite{gomez2022exploring}	& LSTM	& attention	& polyline	& LSTM	& states	& unimodal	\\
Multimodal~\cite{cui2019multimodal}	& raster	& conv	& raster	& conv	& states	& weighted set	\\
MultiPath~\cite{chai2019multipath}	& raster	& conv	& raster	& MLP	& states	& GMM w/ static anchors	\\
MultiPath++~\cite{varadarajan2021multipath++}	& LSTM	& RNNs+maxpool	& polyline	& MLP	& control poly	& GMM	\\
Trajectron++\cite{salzmann2020trajectron++}	& LSTM	& RNNs+attention	& raster	& GRU	& controls	& GMM	\\
CRAT-PRED\cite{schmidt2022crat}	& LSTM	& GNN+attention	& --	& MLP	& states	& weighted set	\\
\midrule					
\textbf{Ours - Social baseline}	& LSTM	& GNN+attention	& --	& LSTM	& states	& weighted samples	\\
\textbf{Ours - Map baseline}	& LSTM	& GNN+attention	& polyline 	& LSTM	& states	& weighted samples	\\
\bottomrule
\end{tabular}
\label{table:related_work}
\end{center}
\end{table*} 

In order to classify DL based MP methods, we distinguish several important features: Motion history, Social information (agent interactions), Map information (road encoding), how the model returns the output trajectory and its corresponding distribution. Table \ref{table:related_work} summarizes several SOTA methods, inspired in the survey proposed by \cite{varadarajan2021multipath++}.

\begin{itemize}

\item \textbf{Motion history}: Most methods encode the sequence of past observed states using 1D-convolution \cite{liang2020learninggraph} \cite{mercat2020multiattentmotion}, able to model spatial information, or via a recurrent net \cite{gomez2022exploring} \cite{alahi2016social} (LSTM, GRU), which are more useful to handle temporal information. Other methods that use a raster version of the whole scenario represent the agent states rendered as a stack of binary mask images depicting agent oriented bounding boxes \cite{gilles2021home}. On the other hand, other approaches encode the past history of the agents in a similar way to the road components of the scene given a set of vectors or polylines \cite{zhao2020tnt, gao2020vectornet} that can model the high-order interactions among all components, or even employing attention to combine features across road elements and agent interactions \cite{ngiam2021scene}.

\item \textbf{Social information}: In complex scenarios, motion history encoding of a particular target agent is not sufficient to represent the latent space of the traffic situation, but the algorithm must deal with a dynamic set of neighbouring agents around the target agent. Common techniques are aggregating neighbour motion history with a permutation-invariant set operator: soft attention \cite{ngiam2021scene, gomez2022exploring}, a combination of soft attention and RNN \cite{varadarajan2021multipath++} / GNN \cite{schmidt2022crat} or social pooling \cite{alahi2016social, gupta2018sgan}. Raster based approaches rely on 2D convolutions \cite{chai2019multipath} \cite{casas2021mp3} over the spatial grid to implicitly capture agent interactions in such a way long-term interactions are dependent on the neural network receptive fields.

\item \textbf{Map information}: High-fidelity maps~\cite{can2022maps} have been widely adopted to provide offline information (also known as physical context) to complement the online information provided by the sensor suite of the vehicle and its corresponding algorithms. Recent learning-based approaches \cite{mahjourian2022occupancy, casas2018intentnet, ivanovic2021heterogeneous}, which present the benefit of having probabilistic interpretations of different behaviour hypotheses, require to build a representation to encode the trajectory and map information. Map information is probably the feature with the clearest dichotomy: raster vs vector treatment. The raster approach encodes the world around the particular target agent as a stack of images (generally from a top-down orthographic view, also known as Bird's Eye View). This world encoding may include from agent state history, agent interactions and usually the road configuration, integrated all this different-sources information as a multi-channel image \cite{gilles2021home}, in such a way the user can use an off-the-shelf Convolutional Neural Network (CNN) based pipeline in order to leverage this powerful information. Nevertheless, this representation has several downsides: constrained field of view, difficulty in modeling long-range interactions and even difficulty in representing continuous physical states due to the inherent world to image (pixel) discretization. On the other hand, the polyline approach may describe curves, such as lanes, boundaries, intersections and crosswalks, as piecewise linear segments, which usually represents a more compact and efficient representation than using CNNs due to the sparse nature of road networks. Some state-of-the-art algorithms not only describe the world around a particular agent as a set-of-polylines \cite{khandelwal2020if} \cite{zhao2020tnt} in an agent-centric coordinate system, but they also leverage the road network connectivity structure \cite{liang2020learninggraph} \cite{zeng2021lanercnn} treating road lanes as a set of nodes (waypoints) and edges (connections between waypoints) in a graph neural network so as to include the topological and semantic information of the map.

\item \textbf{Decoder}: Pioneering works of DL based MP usually adopt the autoencoder architecture, where the decoder is often represented by a recurrent network (GRU, LSTM, etc., specially designed to handle temporal information) to generate future trajectories in an autoregressive way, or by CNNs \cite{gilles2021home} \cite{gilles2021gohome} / MLP \cite{liang2020learninggraph} \cite{schmidt2022crat} using the non-autoregressive strategy. The method may use an autoregressive strategy where the pipeline generates tokens (in this case, positions or relative displacements) in a sequential manner, in such a way the new output is dependent on the previously generated output, whilst MLP \cite{schmidt2022crat}, CNN \cite{gilles2021home} or transformer \cite{ngiam2021scene} based strategies usually follow a non-autoregressive strategy, where from a latent space the whole future trajectory is predicted.

\item \textbf{Output}: The most popular model output representation is a sequence of states (absolute positions) or state differences (relative displacements for any dimension considered). The spacetime trajectory may be intrinsically represented as a continous polynomial representation or a sequence of sample points. Other works \cite{gilles2021home} \cite{gilles2021gohome} first predict a heatmap and then decode the corresponding output trajectories after sampling points from the heatmap, whilst \cite{casas2021mp3} learns a cost function evaluator of trajectories that are enumerated heuristically instead of being generated by a learned model. 

\item \textbf{Trajectory Distribution}: The choice of output trajectory distributions has several approaches on downstream applications. Regardless the agent to be predicted is described as a (non-)holonomic \cite{triggs1993motion} platform, an intrinsic property of the motion prediction problem is that the agent must follow one of a diverse set of possible future trajectories. A popular choice to represent a multimodal prediction are Gaussian Mixture Models (GMMs) due to their compact parameterized form, where mode collapse (associated frequently to GMMs) is addressed through the use of trajectory anchors \cite{chai2019multipath} or training  tricks \cite{cui2019multimodal}. Other approaches model a discrete distribution via a collection of trajectory samples extracted from a latent space and decoded by the model \cite{rhinehart2018r2p2} or over a set of trajectories (fixed or a priori learned) \cite{liang2020learninggraph}.

\end{itemize}

After classifying main SOTA methods, we conclude this section presenting the main characteristics of our baseline approaches. We make use of LSTM to encode the past motion history, GNN in combination with soft-attention the compute social interactions, a set-of-polylines to represent the most important map information and LSTM to decode the trajectories from the latent space. The output multimodal prediction is represented by a set of states with their corresponding confidences indicating the most plausible modes.

\subsection{Datasets}
\label{sec:datasets}

Large-scale annotated datasets have been proposed to impulse the research on the MP task. Focusing on self-driving cars, we find several state-of-the-art datasets with their corresponding benchmarks. Table \ref{table:datasets_comparison} summarizes the comparison of some of the main SOTA datasets for Vehicle Motion Prediction. 
NuScenes Prediction~\cite{caesar2020nuscenes} is a public large-scale dataset with more than 40k scenarios for AD where the trajectories are represented in the x-y BEV coordinate system at 2 Hz. 
Up to 2s of past history may be used so as to predict 6s future trajectories for each agent. 
The Waymo Open Motion Prediction \cite{ettinger2021large} consists of 1.1 M examples (urban and suburban environments) time-windowed from 104k scenarios, each one with a duration of 20s originally sampled at a frequency of 10 Hz. Each scenario consists of 1.1s of past observation (agent dimensions, velocity and acceleration vectors, orientation, turn signal state and angular velocity) and 8s of future states, which are resampled at 5 Hz.

In this work, we focus on the \textbf{Argoverse 1} Motion Forecasting Dataset~\cite{chang2019argoverse}, which is the most frequently used dataset for MP development in the field of AD. It contains more than 300K scenarios from Miami and Pittsburgh, each traffic scenario contains a 2D BEV centroid of unique objects (so, multi-object tracked) at 10 Hz. The task is to predict the future trajectories of a particular target agent in the next 3s, given the past 2s observations in addition to the HD map features. 

\begin{table}[h]

  \caption{\textbf{Comparison between popular benchmarks for Vehicle Motion Prediction}. Track Length and Agents per scene represent average values per scenario. Leaderboard entries values retrieved on January 13, 2023.}
  \label{table:datasets_comparison}
  \centering
  \begin{tabular}{l c c c}
    \toprule	
    & Argoverse~\cite{chang2019argoverse} & Waymo~\cite{ettinger2021large} & NuScenes~\cite{caesar2020nuscenes} \\
    \midrule	
    Scenarios & 324k & 104k & 41k \\
    Unique Tracks & 11.7M & 7.6M & - \\
    Track Length & 2.48 s & 7.04 s & - \\
    Agents per scene & 50 & - & 75 \\
    Cities & 2 & 6 & 2 \\
    Total Time & 320 h & 574 h & 5.5 h \\
    Scenario Duration & 5 s & 9.1 s & 8 s \\
    Prediction Duration & 3 s & 8 s & 6 s \\
    Sampling Rate & 10 Hz & 10 Hz & 2 Hz \\  
    Evaluated categories & 1 & 3 & 1 \\
    Multi-agent evaluation & $\times$ & \checkmark & $\times$ \\
    Vector Map & \checkmark & \checkmark & \checkmark \\
    Leaderboard Entries & 290 & 34 & 36 \\
    \bottomrule	
  \end{tabular}
\end{table}

\section{Our approach}
\label{sec:our_approach}

\begin{figure*}[!ht]
    \centering
    \setlength{\tabcolsep}{2.0pt}
    \includegraphics[width=0.95\linewidth]{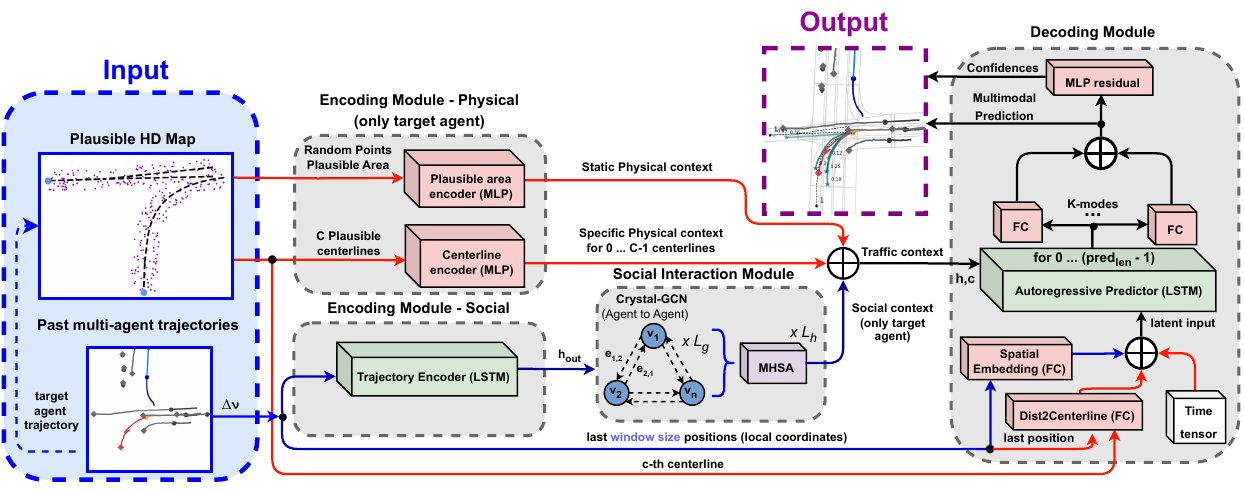}
    \caption{Overview of our final Motion Prediction model (\textbf{\textcolor{blue}{Blue links}} and \textbf{\textcolor{red}{Red links}} represent \textbf{\textcolor{blue}{Social}} and \textbf{\textcolor{red}{Map}} information respectively). We distinguish three main blocks: 1) \textbf{Encoding module}, which uses plausible HD Map information (specific centerlines and driveable area around) and agents past trajectories to compute the motion and physical latent features, 2) \textbf{Social Attention module}, which calculates the interaction among the different agents and returns the most relevant social features, 3) \textbf{Decoding module}, responsible for calculating the multimodal prediction by means of an autoregressive strategy concatenating low-level map features and social features as a baseline, as well as iterating over the different latent centerlines for specific physical information per mode.}
\label{fig:main_diagram}
\end{figure*}

As stated in Section \ref{sec:introduction}, considering the trade-off between curated input data and complexity, we aim to achieve competitive results on the Argoverse Benchmark~\cite{chang2019argoverse} using powerful DL techniques in terms of prediction metrics (minADE, minFDE), including attention mechanisms and GNNs, while reducing the number of parameters of operations with respect to other SOTA methods. In particular, the inputs for our models are: (i) the agent's past trajectories and their corresponding interactions, as the only input for our social baseline. (ii) an extension where we add a simplified representation of the agent's feasible area as an extra input for our map baseline. Therefore, our models do not require full-annotated (including, topological, geometric and semantic information) HD Maps or rasterized BEV representations of the scene to compute the physical context. \\
We use a simple-yet-powerful map preprocessing algorithm where the target agent's trajectory is initially filtered. Next, we compute the feasible area where the target agent can interact taking into account only the geometric information of the HD Map. Fig. \ref{fig:main_diagram} illustrates an overview of our final approach.

\subsection{Social Baseline}
\label{subsec:social_baseline}

Our social baseline is inspired in the architecture proposed by \cite{schmidt2022crat}. It uses as input the past trajectories of the most relevant obstacles as relative displacements to feed the Encoding Module (Fig. \ref{fig:main_diagram}). Then, the social information is computed using a Graph Neural Network (GNN), in particular Crystal-Graph Convolutional Network (\textbf{Crystal-GCN}) layers~\cite{xie2018crystal, schmidt2022crat}, and Multi-Head Self Attention (MHSA)~\cite{vaswani2017attention} to obtain the most relevant agent-agent interactions. Finally, we decode this latent information using an \textbf{autoregressive} strategy where the output at the \textit{i-th} step depends on the previous one for each mode respectively in the Decoding Module. The following sections provide in-depth description of the aforementioned modules.

\subsubsection{Preprocessing - Social}
\label{subsubsec:preprocessing_social}

 Multiple methods \cite{liang2020learninggraph} \cite{schmidt2022crat} consider only the vehicles that are observable at \textit{t=0}. 
 In our case, we consider the agents that have information over the full history horizon $T_h$ = \textit{$T_{obs}$} + \textit{$T_{len}$} (\emph{e.g.} 5s timeframe for Argoverse), reducing the number of agents to be considered in complex traffic scenarios. Instead of using absolute 2D-BEV (\textit{xy} plane), the input for the agent \textit{i} is a series of relative displacements:

\begin{equation}
	\Delta \boldsymbol{\nu}^{t}_i = \boldsymbol{\nu}^{t}_i - \boldsymbol{\nu}^{t-1}_i
\end{equation}

Where $\boldsymbol{\nu}^{t}_i$ represents the state vector (in this case, \textit{xy} position) of the agent \textit{i} at timestamp \textit{t}.

\subsubsection{Encoding module - Social}
\label{subsubsec:encoding_social}

Unlike other methods, we do not limit nor fix the number of agents per sequence. Given the relative displacements of all different agents, a single LSTM is used to compute the temporal information of each agent in the sequence:

\begin{equation}
	out, \mathbf{h_{out}}, \mathbf{c_{out}} = \mathrm{LSTM}(\Delta \boldsymbol{\nu}^{obs_{len}}, \mathbf{h_{in}}, \mathbf{c_{in}})
\end{equation}

The input hidden and cell vectors ($\mathbf{h_{in}}, \mathbf{c_{in}}$) are initialized with a tensor of zeros. $\Delta \boldsymbol{\nu}^{obs_{len}}$ represents the relative displacements over the whole past horizon $obs_{len}$. In order to feed the agent-agent interaction module (Social Attention Module in Fig. \ref{fig:main_diagram}), we take the output hidden vector ($\mathbf{h_{out}}$).

\subsubsection{Social Attention module}
\label{subsubsec:social_attention_module}

After encoding the past history of each vehicle in the sequence, we compute the agent-agent interactions to obtain the most relevant social information of the scene. For this purpose, we construct an interaction graph using Crystal-GCN \cite{xie2018crystal} \cite{schmidt2022crat}. %
Then, MHSA \cite{vaswani2017attention} is applied to enhance the learning of agent-agent interactions. 

Before creating the \textbf{interaction mechanism}, we split the temporal information in the corresponding scenes, taking into account that each traffic scenario may have a different number of agents. The interaction mechanism is defined in \cite{schmidt2022crat} as a bidirectional fully-connected graph, where the initial node features $\mathbf{v}_i^{(0)}$ are represented by the latent temporal information for each vehicle $\mathbf{h}_{i,out}$ computed by the motion history encoder. On the other hand, the edges from node \textit{k} to node \textit{l} is represented as the vector distance ($\mathbf{e}_{k,l}$) between the corresponding agents at t = \textit{$obs_{len}$} in absolute coordinates, where the origin of the sequence ($x=0,y=0$) is represented by the position of the target at t = \textit{$obs_{len}$}:

\begin{equation}
	\mathbf{e}_{k,l} = \boldsymbol{\nu}^{obs_{len}}_k - \boldsymbol{\nu}^{obs_{len}}_l \text{,}
\end{equation}

Given the interaction graph (nodes and edges), the Crystal-GCN, proposed by \cite{xie2018crystal}, is defined as:

\begin{multline}
	\mathbf{v}_i^{(g+1)} = \mathbf{v}_i^{(g)} + \\
	\sum_{j = 0 \textbf{:} j \neq i }^{N} \sigma \left( \mathbf{z}_{i,j}^{(g)} \mathbf{W}_\mathrm{f}^{(g)} + \mathbf{b}_\mathrm{f}^{(g)} \right)
	\odot \mu \left( \mathbf{z}_{i,j}^{(g)} \mathbf{W}_\mathrm{s}^{(g)} + \mathbf{b}_\mathrm{s}^{(g)}  \right) \text{.}
\end{multline}

This operator, in contrast to many other graph convolution operators \cite{zeng2021lanercnn} \cite{liang2020learninggraph}, allows the incorporation of edge features in order to update the node features based on the distance among vehicles (the closer a vehicle is, the more is going to affect to a particular node). As stated by \cite{schmidt2022crat}, we use $L_g = 2$ layers of the GNN ($g \in 0, \dots , L_g$ denotes the corresponding Crystal-GCN layer) with ReLU and batch normalization as non-linearities between the layers. $\sigma$ and $\mu$ are the sigmoid and softplus activation functions respectively. 
Moreover, $\mathbf{z}_{i,j}^{(g)} = ( \mathbf{v}_i^{(g)} || \mathbf{v}_j^{(g)} ||\mathbf{e}_{i,j} )$ corresponds to the concatenation of two node features in the \textit{$g_{th}$} GNN layer and the corresponding edge feature (distance between agents), N represents the total number of agents in the scene and $\mathbf{W}$ and $\mathbf{b}$ the weights and bias of the corresponding layers respectively.

After the interaction graph, each updated node feature $\mathbf{v}_i^{(L_g)}$ contains information about the temporal and social context of the agent \textit{i}. Nevertheless, depending on their current position and past trajectory, an agent may require to pay attention to specific social information. To model this, we make use of a scaled dot-product Multi-Head Self-Attention mechanism \cite{vaswani2017attention} which is applied to the updated node feature matrix $\mathbf{V}^{(L_g)}$ that contains the node features $\mathbf{v}_i^{(L_g)}$ as rows.

Each head $h \in 1,\dots, L_h$ in the MHSA mechanism is defined as:

\begin{equation}
	\mathrm{head}_h = \mathrm{softmax} \left( \frac{\mathbf{V}^{(L_g)}_{Q_h} \mathbf{V}^{(L_g) T}_{K_h}}{\sqrt{d}}  \right) \mathbf{V}^{(L_g)}_{V_h} \text{.}
\end{equation}

where $\mathbf{V}^{(L_g)}_{Q_h}$ (Query), $\mathbf{V}^{(L_g)}_{K_h}$ (Key) and $\mathbf{V}^{(L_g)}_{V_h}$ (Value) represent the \textit{$h_{th}$} head linear projections of the node feature matrix $\mathbf{V}^{(L_g)}$ and $d$ is the normalization factor corresponding to the embedding size of each head. For our purpose, we use $L_h = 4$ as the total number of heads. 
The result of the softmax weights multiplied by the node feature matrix $\mathbf{V}^{(L_g)}_{V_h}$ (Value) is often referred as the attention weight matrix, representing in this particular case pairwise dependencies among vehicles.

Finally, the updated node feature matrix $\mathbf{SATT}$ is computed as the combination of the different attention heads in a single matrix:

\begin{equation}
	\mathbf{SATT} = (\mathrm{head}_1 || \dots || \mathrm{head}_{L_h}) \mathbf{W}_\mathrm{o} + 
	\begin{pmatrix}
	        \mathbf{b}_\mathrm{o}\\
			\vdots \\
			\mathbf{b}_\mathrm{o}
	\end{pmatrix}.
\end{equation}

Where each row of the final social attention matrix $\mathbf{SATT}$ (output of the social attention module, after the GNN and MHSA mechanisms) represents the interaction-aware feature of the agent \textit{i} with surroundings agents, considering the temporal information under the hood, being $\mathbf{W}_\mathrm{o}$ / $\mathbf{b}_\mathrm{o}$ the corresponding weight and bias of the layer that merges the different attention heads. 
Regarding the Argoverse Motion Forecasting benchmark, we \textbf{only consider} the row of the final matrix that takes into account the interactions of the target agent with surrounding obstacles.

\subsection{Map Baseline} 
\label{subsec:map_baseline}

As mentioned before, in this work we extend our social baseline using minimal HD map information, 
from which we discretize the feasible area $\mathcal{P}$ of the target agent as a subset of $r$ randomly sampled points $\{p_0 , p_1 ... p_r\}$ (low-level features) around the plausible centerlines (high-level and well-structured features) considering the velocity and acceleration of the target agent in the last observation frame (Fig. \ref{fig:hdmap_filtering}).
This is a map preprocessing step, therefore the model never sees the HD map (either vectorized or rasterized). 

\subsubsection{Preprocessing - Map} 
\label{subsubsec:preprocessing_map}

Multiple approaches have tried to predict realistic trajectories by means of learning physically feasible areas as heatmaps or probability distributions of the agent’s future location~\cite{dendorfer2020goalgan, sadeghian2019sophie, gilles2021home}. These approaches require either a top-view RGB BEV image of the scene, or a HD map with exhaustive topological, geometric and semantic information (commonly codified as channels). This information is usually encoded using a CNN and fed into the model together with the social agent's information~\cite{dendorfer2020goalgan, sadeghian2019sophie, gao2020vectornet}. 
%
%
As observed, most SOTA methods utilize HD map information to enhance the latent information to decode the future trajectories, but heavy computation or raw data features are required. \textbf{Effectively} and \textbf{efficiently} exploiting HD maps is a must for MP models to produce accurate and plausible trajectories in real-time applications, specially in the field of AD, providing specific map information to each agent based on its kinematic state and geometric HD map information.

\begin{figure*}[]

\centering

\includegraphics[width=0.95\linewidth]{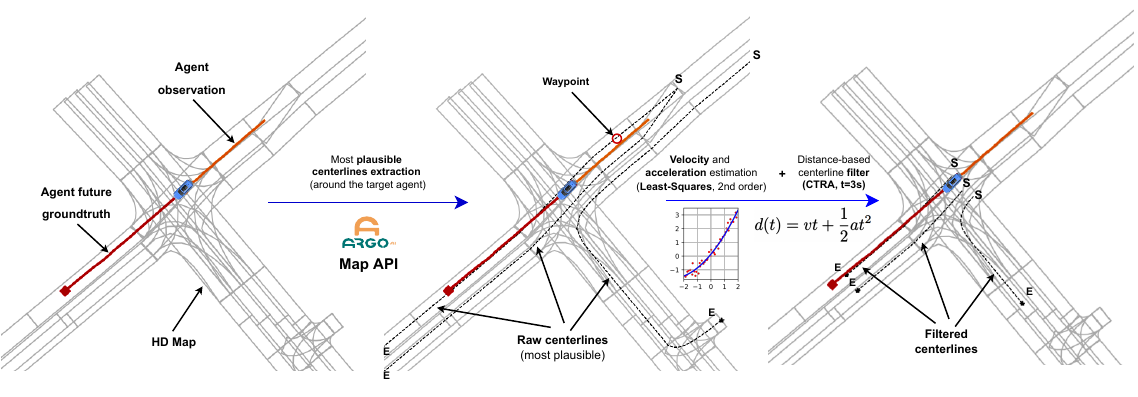}

\caption{Plausible centerlines estimation. Left: General view of the scene, only considering the target agent (\textbf{\textcolor{YellowOrange}{observation (2 s)}} and \textbf{\textcolor{Red}{future ground-truth (3 s)}}) and HD Map around its last observation (position of the \textbf{\textcolor{blue}{blue}} vehicle). Center: \textbf{Centerlines} proposed by the Argoverse Map API (maximum number of centerlines M set to 3). Right: We filter the input observation by means of Least-Squares (2nd order) algorithm to estimate the velocity and acceleration of the agent. Then, the distance considering a CTRA (Constant Turn Rate Acceleration) model and a prediction horizon of 3 s are used computed to obtain the end-points \textbf{E} of the \textbf{final proposals}. Start-points \textbf{S} are the closest centerlines waypoints to the agent in the last observation frame.}
\label{fig:hdmap_filtering}

\end{figure*}

First, in order to obtain the most plausible \textit{M} lane candidates, we make use of the pertinent functions of the Argoverse Map API \cite{chang2019argoverse} as illustrated by \cite{khandelwal2020if} to choose the closest centerlines to the last observation data of the target agent, which are going to represent its most plausible future trajectories. Nevertheless, these lane candidates present various lengths (Fig. \ref{fig:hdmap_filtering}) regardless the past trajectory of the corresponding agent or the ground-truth prediction.

As stated by Argoverse 1 \cite{chang2019argoverse} and Argoverse 2 \cite{wilson2023argoverse}, vehicles are the only evaluated object category as the target agent in the Argoverse 1 Motion Forecasting dataset. Then, considering a vehicle as a rigid structure with non-holonomic \cite{triggs1993motion} features (no abrupt motion changes between consecutive timestamps) and the road driving task is usually described as anisotropic \cite{ross1989planning} (most relevant features are found in a specific direction, in this case the lanes ahead) we \textbf{obtain a simplified version of the HD map}.

Argoverse input trajectory data often presents noise associated to the real-world data capturing process. 
Then, in order to estimate the dynamic variables of the target agent in the last observation frame $t_{obs_{len}}$, first we filter (Fig. \ref{fig:hdmap_filtering}) the target agent past observation
using the Least-Squares algorithm (polynomial order = 2) per axis. By doing so, and assuming the agent is moving with a constant acceleration, we are able to calculate the dynamic features (velocity and acceleration) of the target agent in $t_{obs_{len}}$. Then, a vector of $obs_{len}$ - 1 and $obs_{len}$ - 2 length is computed to estimate the velocity and acceleration respectively as $V_{i}=\frac{X_{i}-X_{i-1}}{t_{i}-t_{i-1}}$ and $A_{i}=\frac{V_{i}-V_{i-1}}{t_{i}-t_{i-1}}$, where $X_{i}={[x_{i},y_{i}]}$ represents the 2D position of the agent at each observed frame $i$ as state above.
Furthermore, we summarize these vectors as scalars by obtaining a smooth estimation assigning less importance (higher forgetting factor $\lambda$) to the first observations, in such a way most recent observations are the key to determine the current kinematic state of the agent:

\begin{equation}
    \hat{\psi}_{obs_{len}} = \sum_{t=0}^{obs_{len}}\lambda^{obs_{len} - t}\psi_t
    \label{eq:dynamic_feats_last_observation_frame}
\end{equation}

where $obs_{len}$ is the number of observed frames, $\psi_t$ is the estimated velocity/acceleration at the $t_{i}$ frame, $\lambda\in(0, 1)$ is the forgetting factor, and $\hat{\psi}_{t_obs}$ is the smoothed velocity/acceleration estimation at the last observation frame. Once the kinematic state is computed, we estimate the travelled distance (Fig. \ref{fig:hdmap_filtering}) assuming a \textbf{CTRA} (Constant Turn Rate Acceleration) physics-based model at any timestamp \textit{t}: 

\begin{equation}
    d(t) = vt + \frac{1}{2}at^2
\end{equation}

 Assuming non-holonomic constraints \cite{triggs1993motion}, which are inherent of standard road vehicles, 
 these should follow a smooth trajectory in a short-mid term prediction.
 Then, we process these raw candidates into \textbf{plausible lane trajectories} so as to use them as plausible physical information. First, we find the closest waypoint to the last observation data of the target agent, which will represent the start point of the plausible centerline. Second, we evaluate the above-mentioned travelled distance along the raw centerlines. We determine the end-point index $p$ of the centerline $m$ as the waypoint (each discrete node of the centerline) where the accumulated distance (considering the $\mathcal{L}_2$ distance between each waypoint) is greater or equal than the above-computed $d(obs_{len})$:

\begin{equation}
    p \quad \textbf{:} \quad d(obs_{len}) \leq \sum_{p=start_{point}}^{centerline_{length}} \mathcal{L}_2(w(p+1),w(p))
\end{equation}
 
\begin{figure*}[]
 \centering
\subfigure[]{\label{subfig:hdmap_filtered_example_1}\includegraphics[width=0.48\textwidth]{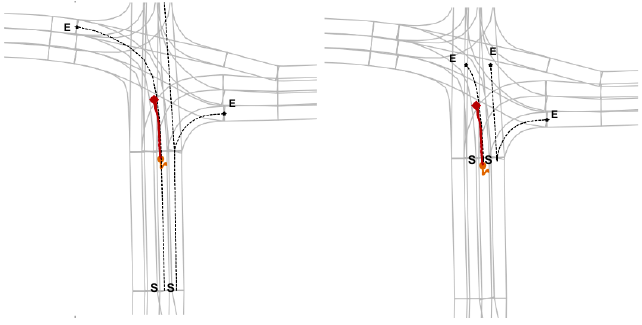}}
~ 
\subfigure[]{\label{subfig:hdmap_filtered_example_2}\includegraphics[width=0.48\textwidth]{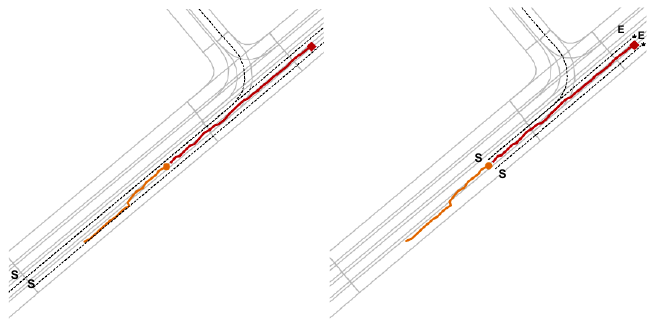}}

 \caption{Some challenging examples of our preprocessing step to obtain relevant map features. In both scenarios the target agent (\textbf{\textcolor{YellowOrange}{observation (2 s)}} and \textbf{\textcolor{Red}{future ground-truth (3 s)}}) presents a noticeable noisy past trajectory and the provided raw \textbf{centerlines} do not consider the current kinematic state of the vehicle. (a) The agent is stopped (maybe due to an stop, pedestrian crossing or red traffic light use case). We estimate a minimum travelled distance of 25 m in these situations to determine the centerline end-points \textbf{E}. (b) In this scenario, we can observe how the raw centerlines consider way more distance (both ahead and behind) than required. Our kinematic-based filter is able to minimize these proposals in an interpretable way to serve as prior information to the MP model
 }
 \label{fig:hdmap_filtered_examples}

\end{figure*}

 Finally, we perform a cubic interpolation between the start-point and the end-point of the corresponding centerline $m$ in order to have the same steps than the prediction horizon $pred_{len}$. Fig. \ref{fig:hdmap_filtering} summarizes this HDMap filtering process in order to have plausible centerlines in an efficient and interpretable way. Moreover, Fig. \ref{fig:hdmap_filtered_examples} illustrates how after our filtering process the end-points of the plausible centerlines are noticeable closer to the ground-truth prediction at the final timestep. Table \ref{table:error_goals} illustrates different preprocessing methods to filter the raw centerlines proposed by the Argoverse Map API. We compare our proposed method, Constant Turn Rate Acceleration with Least-Squares (2nd order) past trajectory filtering, against other approaches, such as only considering the velocity of the target agent in $t_{obs_{len}}$ (instead of velocity and acceleration), as well as compute the kinematic state of the vehicle without filtering the input trajectory. As expected, the best prior information, considering the mean and median $\mathcal{L}_2$ distance over the whole validation split between the target agent's ground-truth end-point and filtered centerlines end-points, is obtained when considering the velocity and acceleration in the kinematic state and filtering the input with the least-squares method. 
 
 In addition to these \textbf{high-level} and well-structured centerlines, we apply point location perturbations (Fig. \ref{fig:results}, second column) to all plausible centerlines under a $\mathcal{N}(0, 0.2)$ [m] distribution~\cite{ye2021tpcn} in order to discretize the plausible area $\mathcal{P}$ as a subset of $r$ randomly sampled points $\{p_0 , p_1 ... p_r\}$ (\textbf{low-level} features) around the plausible centerlines. By doing this, we  may have a common representation of the plausible area, defined as low-level map features. We make use of a normal distribution $\mathcal{N}$ to calculate these random points as an additional regularization term instead of using the lanes boundaries in order to prevent overfitting in the encoding module, in a similar way that data augmentation is applied to the past trajectories. 

 \begin{table}[t]
    \centering
    \caption{\textbf{Error comparison} (in terms of $\mathcal{L}_2$~distance) between the target agent ground-truth end-point and proposed centerlines end-points with different preprocessing methods in the validation split (39,472 samples). CTRV and CTRA stand for \textit{Constant Turn Rate Velocity} and \textit{Constant Turn Rate Acceleration} respectively. We represent both the median and median values for each case.
    }
    \resizebox{\linewidth}{!}{
    \begin{tabular}{l c c}
        \toprule
         Preprocessing method & $\mathcal{L}_2$ Mean~(m) $\downarrow$ & $\mathcal{L}_2$ Median~(m) $\downarrow$ \\
         \midrule
         CTRV without past trajectory filtering & 6.22 & 4.33 \\
         CTRA without past trajectory filtering & 8.78 & 6.31 \\
         CTRV + Least-Squares (2nd order) & 6.14 & 4.21 \\
         CTRA + Least-Squares (2nd order) & \textbf{5.37} & \textbf{3.89} \\
         \bottomrule
    \end{tabular}}
    \label{table:error_goals}
\end{table}

\subsubsection{Encoding module - Map}
\label{subsubsec:encoding_map}

In order to calculate the latent map information, we employ a plausible area and centerline encoder (Fig. \ref{fig:main_diagram}) to process the low-level and high-level map features respectively. Each of these encoders are represented by a Multi-Layer Perceptron (MLP). First, we flat the information along the points dimension, alternating the x-axis and y-axis information. Then, the corresponding MLP (three layers, with batch normalization, interspered ReLUs and dropout in the first layer) transforms the interpretable absolute coordinates around the origin ($x=0, y=0$) into representative latent physical information. The static physical context (output from the plausible area encoder) will serve as a common latent representation for the different modes, whilst the specific physical context will illustrate specific map information for each mode.

\subsection{Decoding module}

The decoding module is the third component of our baselines, as observed in Fig. \ref{fig:main_diagram}. The decoding module consists of an \textbf{LSTM} network, which recursively estimate the relative displacements for the future timesteps, in the same way we studied the past relative displacements in the Motion History encoder. Regarding the social baseline, the model uses the social context computed by the Social Interaction Module, only paying attention to the target agent row. Then, only the social context corresponds to the whole \textit{traffic context} of the scenario, representing the input hidden vector of the autoregressive LSTM predictor. On the other hand, in terms of the map baseline, for a mode \textit{m}, we identify the latent \textit{traffic context} as the concatenation of the social context, static physical context and specific physical context as stated in Sec. \ref{subsubsec:encoding_map}, which will serve as input hidden vector $\mathbf{h}$ of the LSTM decoder. In both cases (social and map baselines), the cell vector $\mathbf{c}$ is initialized with a vector of zeros of the same dimension. \\

Regarding the LSTM input, in the social case it is represented by the encoded past $n$ relative displacements of the target agent after a spatial embedding, whilst the map baseline adds the encoded vector distance between the current absolute target position and the current centerline, as well as the current scalar timestamp $t$, as illustrated in Fig. \ref{fig:main_diagram}. In both cases (social and map baselines), we process the output of the LSTM using a standard Fully-Connected (FC) layer (one per mode). Once we have the relative prediction in the timestep $t$, we shift the initial past observation data in such a way we introduce our last-computed relative displacement at the end of the vector, removing the first data. We identify this technique as a \textit{temporal decoder}, where a window of size $n$ is analized by the autoregressive decoder in contrast to other techniques \cite{dendorfer2020goalgan} \cite{sadeghian2019sophie} \cite{gupta2018social} where only the last data is considered. Finally, after performing relative displacements to absolute coordinates operation, we obtain our multimodal predictions $\hat{Y} \in \mathbb{R}^{k \times pred_{len} \times data_{dim}}$, where $k = 6$ represents the number of modes, $pred_{len} = 30$ represents the prediction horizon and $data_{dim}$ represents the data dimensionality, in this case 2 ($xy$, predictions from the BEV perspective). Once the multimodal predictions are computed, they are concatenated and processed by a residual MLP to obtain the confidences (the higher the confidence, the most probable the mode must be, and closer to the ground-truth).

\section{Experiments}

\subsection{Experimental Setup}
\label{sec:experiments}

\paragraph{Dataset} 

As explained in Section~\ref{sec:datasets} we use the public available Argoverse 1 Motion Forecasting Dataset~\cite{chang2019argoverse} which consists of 205942 training samples, 39472 validation samples and 78143 test samples. 
%
For training and validation, full 5-second trajectories are provided, while for testing, only the first 2 seconds trajectories are given. 
%

\paragraph{Metrics}

We evaluate the performance of our models using the standard metrics for multimodal MP \cite{chang2019argoverse}: (i) Average Displacement Error (\textbf{ADE}), which averages the $L_2$ distances between the ground truth and predicted output across all timesteps, (ii) Final Displacement Error (\textbf{FDE}), which computes the $L_2$ distance between the final points of the ground truth and the predicted final position. When the output is multimodal, we generate $k$ outputs (also known as modes) per prediction step and report the metrics for the best out of $k$ outputs, regarding the agent $i$.

We report results for $k=1$ (unimodal case, only the mode with the best confidence is considered) and $k=6$ as this is the standard in the Argoverse Motion Forecasting dataset in order to compare with other models.

\subsection{Implementation Details}

We train our models to convergence using a single NVIDIA \textbf{RTX 3090}, and validate our results on the official Argoverse validation set~\cite{chang2019argoverse}. We use Adam optimizer with learning rate $0.001$ and default parameters, batch size $1024$ and linear LR Scheduler with factor $0.5$ decay on plateaus. We rotate the whole scene regarding the orientation in the last observation frame of the target agent to align this agent with the positive y-axis. The hidden dimension for the Motion History encoder is 64, where both the hidden state $\mathbf{h_{in}}$ and cell state $\mathbf{c_{in}}$ are initialized with zeros (dim = $128$), whilst the MLP encoder for both the specific centerline and plausible area is 128. Regarding the Social Interaction module, the latent vector of the Crystal-GCN layers is 128 and the number of heads in the MHSA module is $L_h = 4$. In terms of the Autoregressive predictor, the spatial embedding and \textit{dist2centerline} modules encode the past data and distance to the specific centerline using a \textit{window size} of 20. We set the number of plausible centerlines (\textit{M}) as 3, which cover most cases (if less than 3 plausible centerlines are available, we add padded centerlines as vector of zeros). The time tensor is a single number that represents the current timestep, in such a way the LSTM input is \textit{(2 $\times$ window size) + 1 = 41}. The regression head is represented by \textit{k=6} FC layers that map the output latent vector returned by the LSTM to the final output relative displacements (dim = 2, xy). Multimodal predictions are processed by an MLP residual of sizes 60, 60 and 6 with interspersed ReLU activations in order to obtain the corresponding confidences.

\paragraph{Augmentations} 

(i) Dropout and swapping random points from the past trajectory, (ii) point location perturbations under a $\mathcal{N}(0, 0.2)$ [m] noise distribution~\cite{ye2021tpcn}. We also apply the well-known hard-mining technique to improve the model's generalization under difficult scenarios. To perform this technique, once we have the social and map baselines, we perform inference on the training set to find the most difficult scenes in terms of minADE. Then, we mine those scenes such that the baselines models perform poorly, and increase their proportion in the batch during training.

\paragraph{Loss}

We use the standard \textbf{Negative Log-Likelihood} (NLL) loss to train our social and map baselines in order to compare the ground-truth points $Y \in \mathbb{R}^{pred_{len} \times data_{dim}} = \{(x_0,y_0) ... (x_{pred_{len}}, y_{pred_{len}})\}$ with our multimodal predictions ($\hat{Y} \in \mathbb{R}^{k \times pred_{len} \times data_{dim}}$), given $k$ modalities (hypotheses) $\mathbf{p}=\{(\hat{x}^1_0,\hat{y}^1_0) ... (\hat{x}^k_{pred_{len}}, \hat{y}^k_{pred_{len}})\}$, with their corresponding confidences $\mathbf{c}=\{c_1 ... c_k\}$ using the following equation:

\begin{equation}
    \text{NLL} = -\log \sum_{k} e^{ \log{c^k} - \frac{1}{2} \sum_{t=0}^{pred_{len}} (\hat{x}^k_t - x_t)^2 + (\hat{y}^k_t - y_t )^2 }
\label{eq:nll}
\end{equation}

Similar to ~\cite{mercat2020multiattentmotion}, we assume the ground-truth points to be modeled by a mixture of multi-dimensional independent Normal distributions over time (predictions with unit covariance). Minimizing the NLL loss maximizes the likelihood of the data for the forecast. Nevertheless, the NLL loss tends to overfit most predictions in a similar direction. As stated above, in the motion prediction task, specially in the Autonomous Vehicles field, we must build a model that not only reasons multimodal predictions in terms of different maneuvers (keep straight, turn right, lane change, etc.) but also different velocity profiles (constant velocity, acceleration, etc.) regarding the same maneuver. For this reason, after the baselines models have been trained, as stated by \cite{kim2022improving}, we add as regularization the Hinge (a.k.a. max-margin) and \textbf{Winner-Takes-All} (WTA) losses to improve the confidences and regressions respectively. Algorithm \ref{alg:1} illustrates how we compute the max-margin and WTA losses. First, we determine the closest mode $m^{*}$ to the ground-truth using the $\mathcal{L}_2$ distance, only considering the end-points. Then, WTA loss is computed using Smooth~$\mathcal{L}_1$ distance taking into account in this case the whole prediction horizon between the best mode and ground-truth prediction. Finally, we apply \textbf{max-margin} loss (a.k.a. Hinge loss) \cite{liang2020learninggraph} \cite{kim2022improving} regarding the confidence of the best mode and a margin ($\epsilon$) of 0.0001. 

\begin{algorithm}[t]
    \caption{Additional regularization: Hinge and WTA loss} 
    \begin{algorithmic}[1]
    
    \STATE\textbf{input}: ground-truth trajectory $(Y \in \mathbb{R}^{pred_{len} \times data_{dim}})$ and output trajectories ($\hat{Y} \in \mathbb{R}^{k \times pred_{len} \times data_{dim}}$), where $k$, $pred_{len}$ and $data_{dim}$ denote the number of modes, prediction horizon and data dimensionality for the target agent.
    
    \STATE\textbf{output} classification loss $\mathcal{L}_{Hinge}$ and regression loss $\mathcal{L}_{WTA}$
    
        \FOR {$m$ in $\{1, 2, \ldots, k\}$}
    	    \STATE $ d_{wta}^m \gets$ Euclidean distance between $\hat{Y}^{m}_{pred_{len}}$ and $Y_{pred_{len}}$
        \ENDFOR
        \STATE $m^* = \arg\min_{m} d_{wta}^{m}$
        \STATE $\mathcal{L}_{reg,WTA} \gets $ Smooth~$\mathcal{L}_1$ loss between $\hat{Y}^{m^*}$ and $Y$
        \STATE $\mathcal{L}_{class,Hinge} = \frac{1}{(K-1)}\sum_{m=1 \setminus m \neq m^*}^{K} \max( 0, c_{k} + \epsilon - c_{m^*})$

        \STATE \textbf{return} $\mathcal{L}_{WTA}$, $\mathcal{L}_{Hinge}$
        
    \end{algorithmic} 
\label{alg:1}
\end{algorithm}

Therefore, our loss function is:

\begin{equation}
    \mathcal{L} = \alpha \mathcal{L}_{NLL} + \beta \mathcal{L}_{Hinge} + \gamma \mathcal{L}_{WTA}
\label{eq:loss}
\end{equation}

Where $\alpha=1.0$ , $\beta=0.1$ and $\gamma=0.65$ initially, and can be manually adjusted during training (especially $\gamma$).

\subsection{Results}
\label{sec:results}

\begin{table*}[thpb]
	\caption{\textbf{Ablation Study for map-free MP on the Argoverse~\cite{chang2019argoverse} validation set}. Our methods are indicated with $\dag$, our highlighted method indicates our map-free baseline (Best Social Model = BSM). Prediction metrics (minADE, minFDE) are reported in meters.}
	\label{table:results_val_social}
	\setlength{\tabcolsep}{5pt}
	\centering
	\begin{tabularx}{\textwidth}{lccccc}
		\toprule
		\multirow{2}{*}{Method} & \multirow{2}{*}{Number of Parameters} &
        \multicolumn{2}{c}{$k=1$} & \multicolumn{2}{c}{$k=6$} \\ 
		& & minADE & minFDE & minADE & minFDE  \\ 
        \midrule
		TPCN \cite{ye2021tpcn} & - & 1.42 & 3.08 & 0.82 & 1.32 \\
        LaneGCN  \cite{liang2020learninggraph} (w/o map) & $\approx 1 M $ & 1.58 & 3.61 & 0.79 & \textbf{1.29} \\
        WIMP \cite{khandelwal2020if} (w/o map) & $> 20 M$ & $1.61$ & 5.05 & 0.86 & 1.39 \\
		CRAT-Pred (LSTM + GNN + Lin. Residual) \cite{schmidt2022crat} & 449K & 1.44 & 3.17 & 0.86 & 1.47 \\
		CRAT-Pred (LSTM + GNN + Multi-Head Self-Attention + Lin. Residual) \cite{schmidt2022crat} & 515K & \textbf{1.41} & \textbf{3.10} & 0.85 & 1.44 \\ 
        \midrule
        $\dag$ LSTM-128 + GNN + MHSA (Baseline social) &  351K  & 1.82 & 3.72 & 0.87 & 1.63 \\
        $\dag$ LSTM-64 + GNN + MHSA &  \textbf{97K}  & 1.77 & 3.68 & 0.86 & 1.61 \\
        $\dag$ LSTM-128 + GNN + MHSA + Lin. Residual &  552K  & 2.02 & 4.16 & 1.02 & 1.95 \\
        $\dag$ LSTM-128 (TDec) + GNN + MHSA  &  365K  & 1.81 & 4.04 & 0.83 & 1.57 \\
        $\dag$ LSTM-64 (TDec) + GNN + MHSA \quad (Best Social Model)  &  105K  & 1.79 & 4.01 & 0.81 & 1.56 \\
        \midrule
        $\dag$ Best Social Model + HardM (10 \%) &  105K  & 1.76 & 3.97 & 0.80 & 1.53 \\
        \rowcolor{gray}$\dag$ Best Social Model + HardM (10 \%) \emph{w/} Loss Hinge + WTA &  105K  & 1.62 & 3.57 & \textbf{0.76} & 1.43 \\
        \bottomrule
	\end{tabularx}
\end{table*}
\begin{table*}[]
	\caption{\textbf{Ablation Study for map-based motion forecasting on the Argoverse~\cite{chang2019argoverse} validation set}. Our methods are indicated with $\dag$. We highlight our map-based baseline method, as a reference for future comparisons.}
	\label{table:results_val_map}
	\setlength{\tabcolsep}{5pt}
	\centering
	\begin{tabularx}{\textwidth}{lccccc}
		\toprule
		\multirow{2}{*}{Method} & \multirow{2}{*}{Number of Parameters} &
        \multicolumn{2}{c}{$k=1$} & \multicolumn{2}{c}{$k=6$} \\ 
		&               & minADE & minFDE & minADE & minFDE  \\ 
        \midrule
		\textbf{$\dag$ Our Map-free Baseline (BSM, No Hard-mining, Loss = NLL)} &  105K  & 1.79 & 4.01 & 0.81 & 1.56 \\
        LaneGCN  \cite{liang2020learninggraph} & 3.7M & \textbf{1.35} & \textbf{2.97} & \textbf{0.71} & \textbf{1.08} \\
        WIMP \cite{khandelwal2020if} (w/o map, NLL loss) & $> 25 M$ & 1.41 & 6.38 & 1.07 & 1.61 \\
        WIMP \cite{khandelwal2020if} (w/o map, EWTA loss) & $> 25 M$ & 1.45 & 3.19 & 0.75 & 1.14 \\
        \midrule
        $\dag$ BSM + Oracle &  \textbf{277K}  & 1.62 & 3.56 & 0.77 & 1.42 \\
        $\dag$ BSM + centerlines=3 &  307K  & 1.60 & 3.53 & 0.76 & 1.39 \\
        $\dag$ BSM + centerlines=3 (1D-CNN) &  432K  & 1.63 & 3.59 & 0.78 & 1.43 \\
        $\dag$ BSM + centerlines=3 loop &  326K  & 1.62 & 3.41 & 0.76 & 1.40 \\
        $\dag$ BSM + centerlines=3 loop + Feasible area &  458K  & 1.62 & 3.40 & 0.76 & 1.40 \\
        $\dag$ BSM + centerlines=3 loop + Feasible area + Dist2Centerline (Best Global Model) &  459K  & 1.61 & 3.40 & 0.75 & 1.39 \\
        \midrule
        $\dag$ Best Global model + HardM (10 \%)  & 459K  & 1.55 & 3.31 & 0.75 & 1.36 \\
        \rowcolor{gray}$\dag$ Best Global model + HardM (10 \%) \emph{w/} Loss Hinge + WTA &  459K  & 1.46 & 3.22 & 0.72 & 1.28 \\
        \bottomrule
	\end{tabularx}
\end{table*}

\paragraph{Ablation studies}

As we state in Section~\ref{sec:introduction} and~\ref{sec:our_approach}, our main goal is to achieve competitive results while being efficient in terms of model complexity, in particular in terms of \textbf{FLOPs} (Floating-Point Operations per second) and \textbf{parameters} in order to enable these models for real-time operation. For this reason, we have proposed light-weight models, whose main input is the history of past trajectories of the agents, complemented by interpretable map-based features. 

In this section we analyze our results and ablation studies, and prove the benefits of our approach for self-driving MP. Table \ref{table:results_val_social} and \ref{table:results_val_map} illustrate our ablation study regarding our social and map baselines respectively. First, we compare our social model with other SOTA models \cite{liang2020learninggraph} \cite{khandelwal2020if} \cite{schmidt2022crat} without map information or with the corresponding module disabled. Our social baseline, trained with NLL loss, presents a number of 351K parameters and 0.87 / 1.63 for minADE and minFDE (k=6) respectively. \\

We perform the following \textbf{ablation studies} in Table \ref{table:results_val_social}: Reduce social hidden dim (including LSTM, GNN and MHSA modules) from 128 to 64, replace the standard head with residual head, replace only last data (standard autoregressive decoder input) with last $N$ data (temporal decoder). We obtain better results with hidden dim = 64, decreasing the number of parameters. Linear residual, standard in most MP models, presents worse results with a much higher number of parameters, since most works use it in a non-autoregressive way, decoding directly from the latent space. On the other hand, using temporal decoder instead of only the last position as LSTM input achieves better results with a slightly higher number of parameters. Then, we conclude our Best Social Model (BSM), as a preliminary stage before implementing the map features, presents the following modifications: social hidden dim = 64 and temporal decoder. Hard-mining and additional losses (Hinge and WTA) applied to the best social model achieve the best social results (Social Baseline).

To integrate the map features (Table \ref{table:results_val_map}) we start from the BSM without hard-mining and with the initial loss (NLL), in order to check how implementing these additional regularization terms help the model to generalize better in both experiments (only social and social+map). We perform the following ablations: compute the most plausible centerline (\textit{M}=1) returned by the \textbf{Argoverse API}, consider \textit{M}=3 centerlines, replace MLP encoder with 1D-CNN encoder in a similar way \cite{mercat2020multiattentmotion}, explicitly iterate over all centerlines as \textbf{specific} deep physical context instead of decoding from a common latent space, add low-level features (feasible area) as a common \textbf{static} deep physical context for each iteration and finally adding an additional component to the LSTM input determined by the vector distance between the considered input window (last $N$ data) and the corresponding centerline. It can be observed that introducing map features increases the number of parameters in exchange of a noticeable metrics decrement, specially in terms of minFDE ($k$ = 6). Considering \textit{M}=3 centerlines instead of only the most plausible centerline allows the model to compute a more diverse set of predictions, while replacing a standard MLP encoder with 1D-CNN encoder increases the number of parameters achieving worse metrics, according to this experimental setup. 

\begin{figure*}[!ht]
    \centering
    \setlength{\tabcolsep}{2.0pt}
    \begin{tabular}{cccc}
    \fbox{\includegraphics[width=0.22\linewidth]{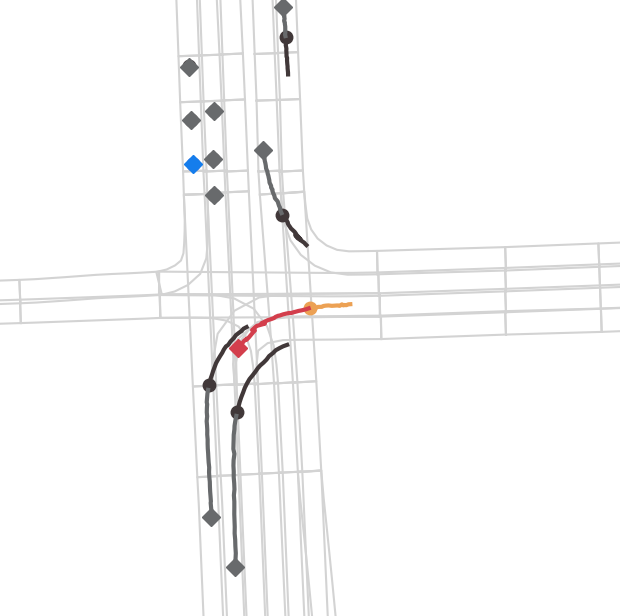}} & 
    \fbox{\includegraphics[width=0.22\linewidth]{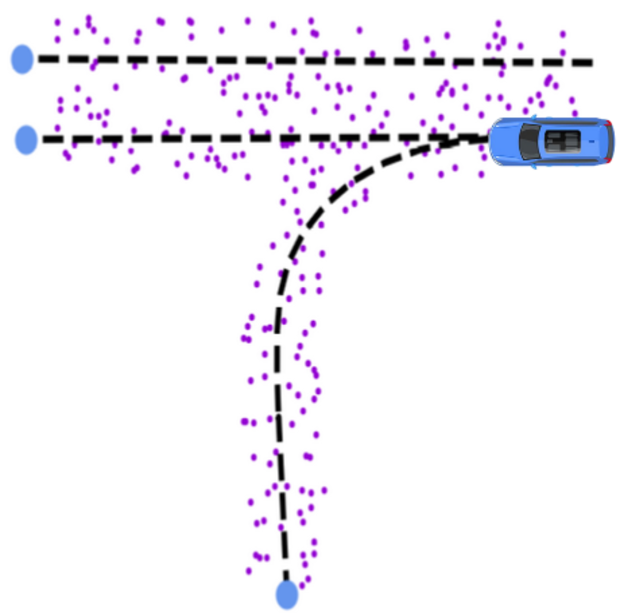}} &
    \fbox{\includegraphics[width=0.22\linewidth]{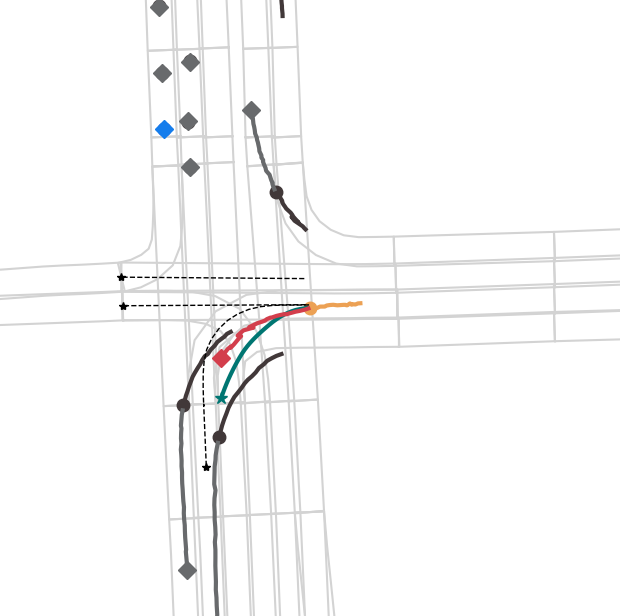}} & 
    \fbox{\includegraphics[width=0.22\linewidth]{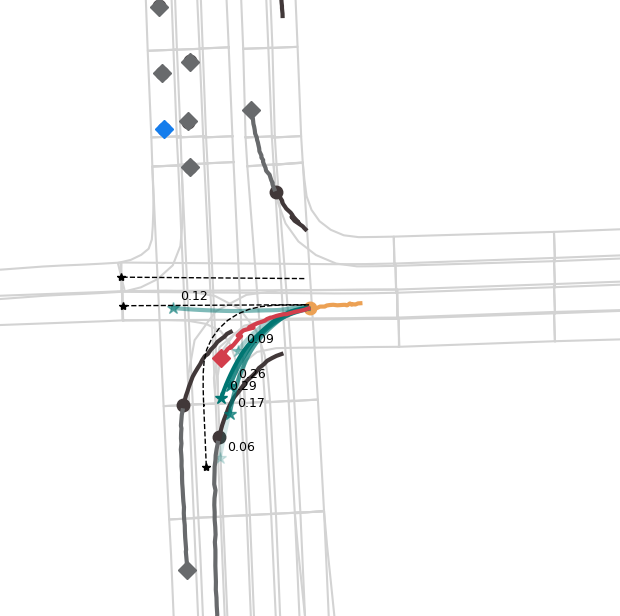}}
    
    \tabularnewline
    
    \fbox{\includegraphics[width=0.22\linewidth]{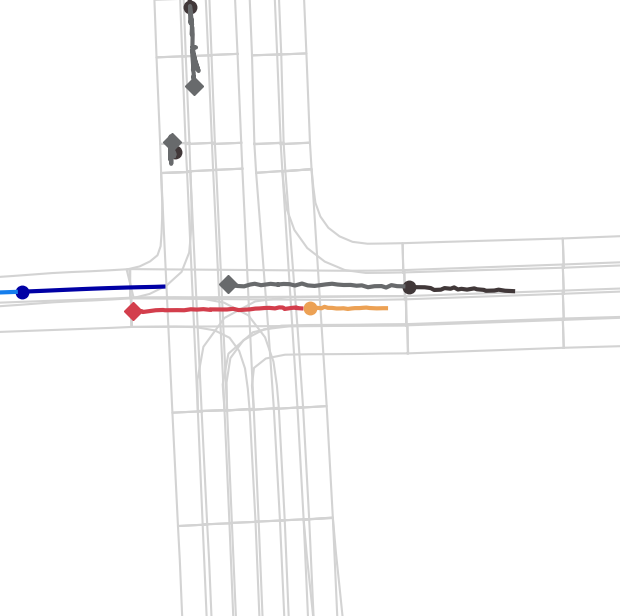}} & 
    \fbox{\includegraphics[width=0.22\linewidth]{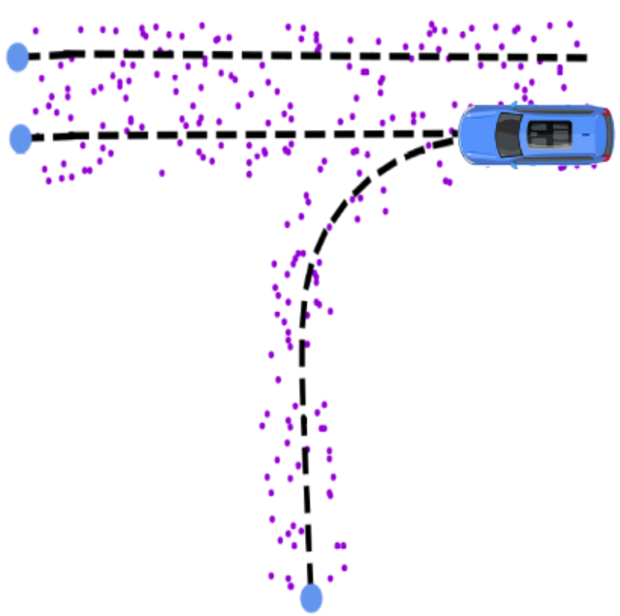}} &
    \fbox{\includegraphics[width=0.22\linewidth]{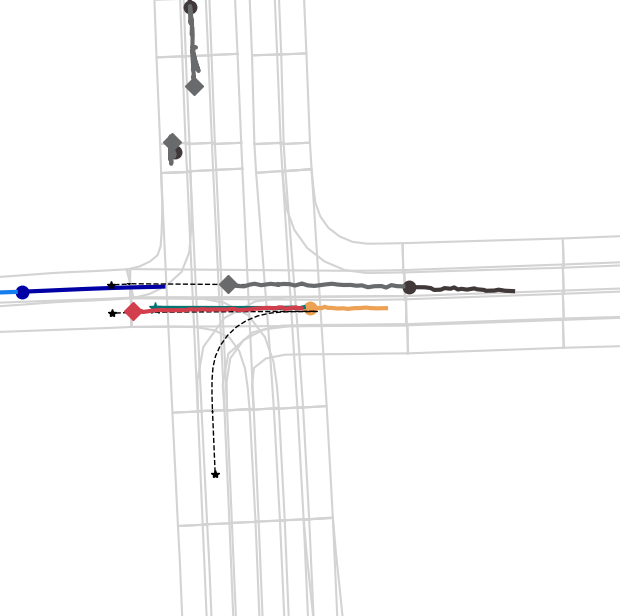}} & 
    \fbox{\includegraphics[width=0.22\linewidth]{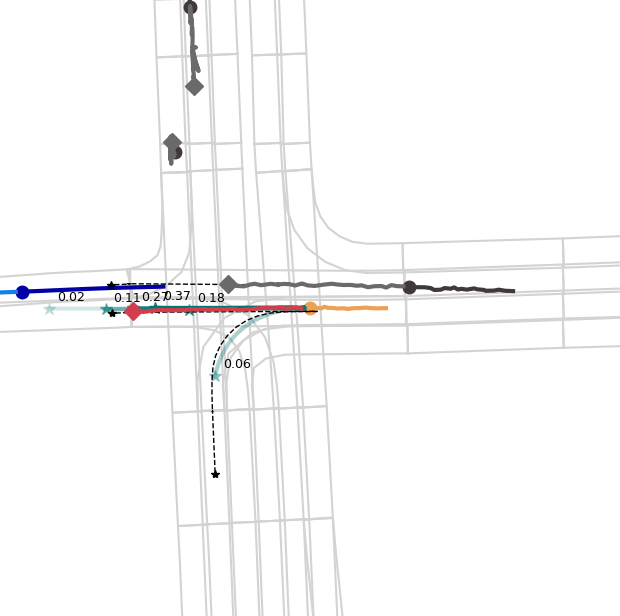}}
    
    \tabularnewline
    
    \fbox{\includegraphics[width=0.22\linewidth]{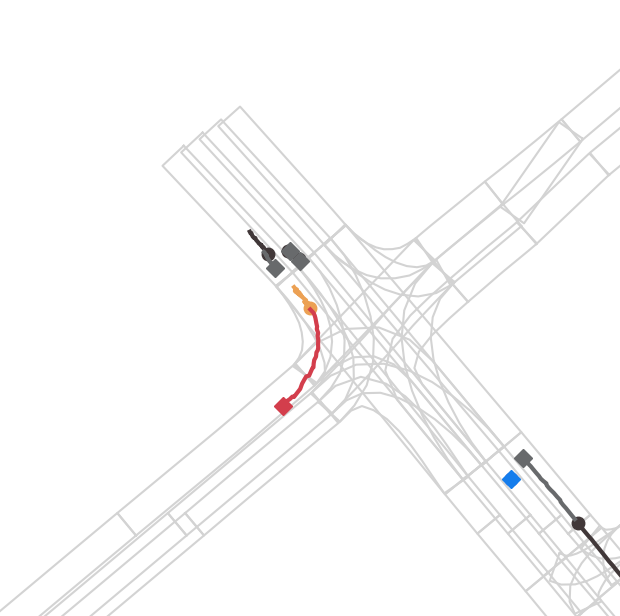}} & 
    \fbox{\includegraphics[width=0.22\linewidth]{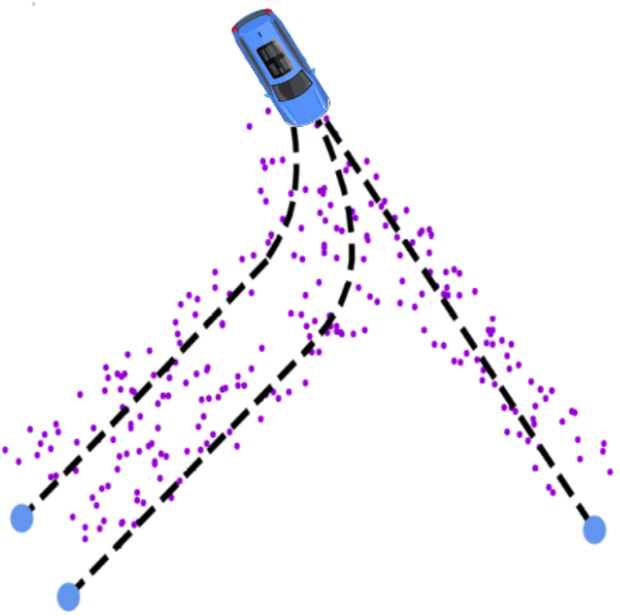}} &
    \fbox{\includegraphics[width=0.22\linewidth]{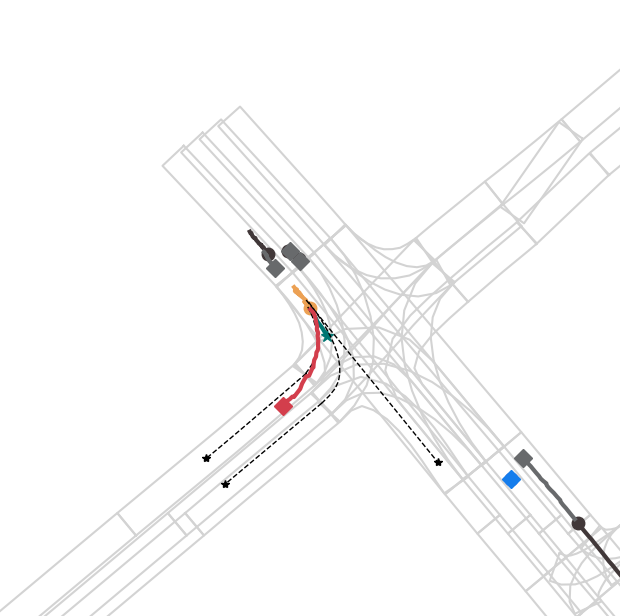}} & 
    \fbox{\includegraphics[width=0.22\linewidth]{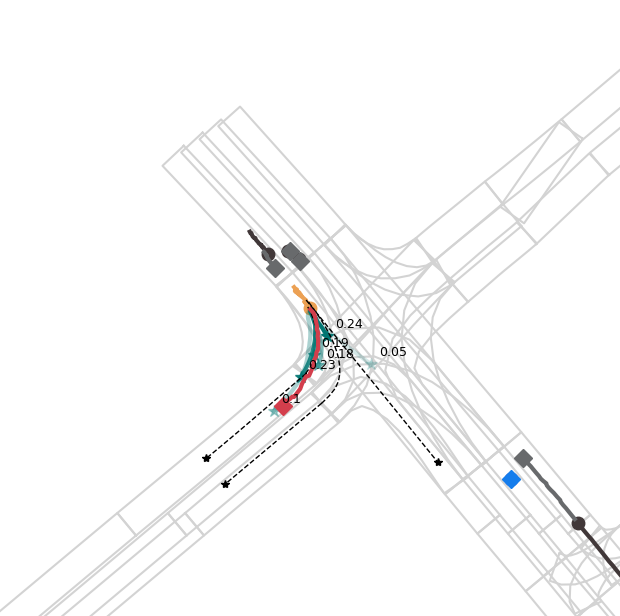}}

    \tabularnewline

    \fbox{\includegraphics[width=0.22\linewidth]{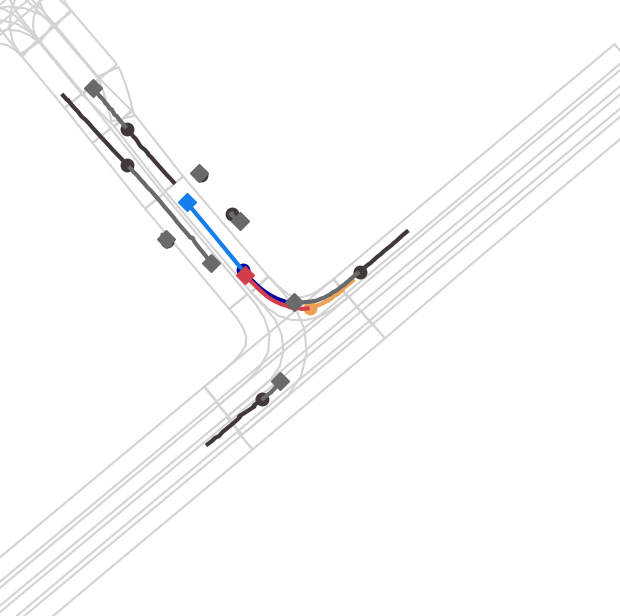}} & 
    \fbox{\includegraphics[width=0.22\linewidth]{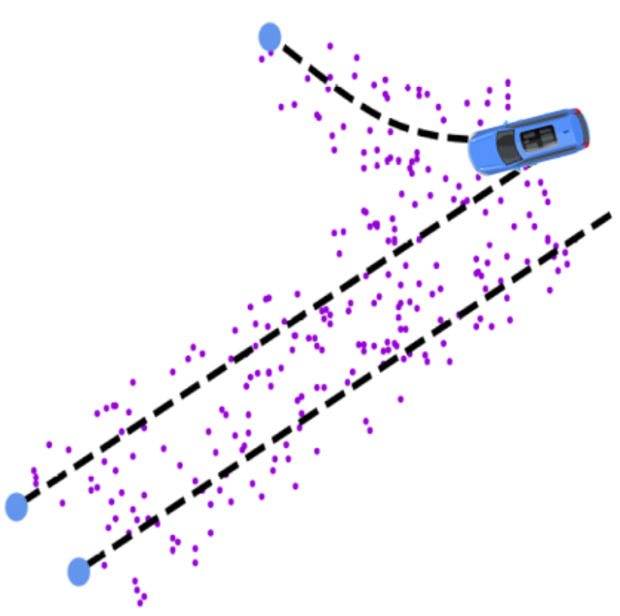}} &
    \fbox{\includegraphics[width=0.22\linewidth]{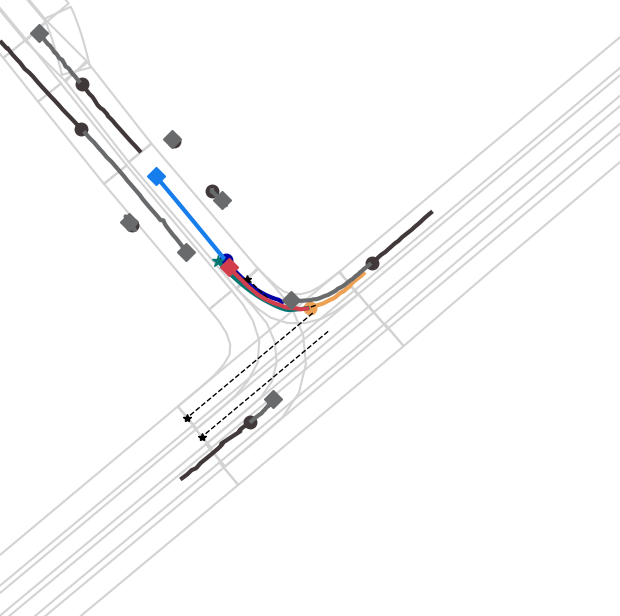}} & 
    \fbox{\includegraphics[width=0.22\linewidth]{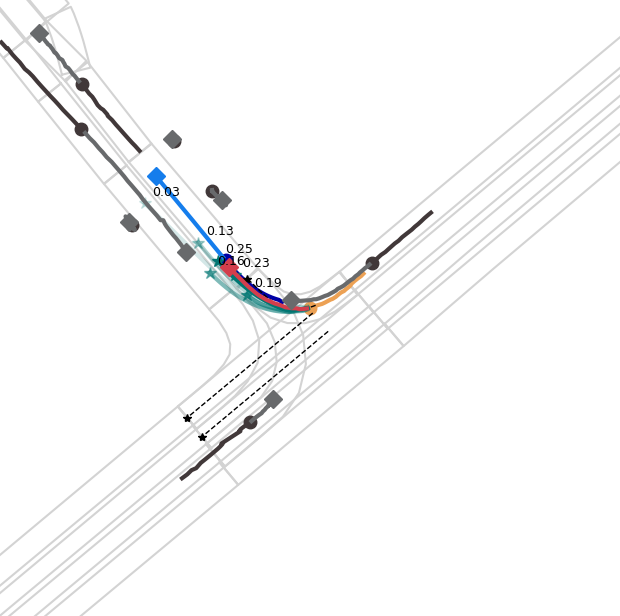}}
    
    \tabularnewline
    
    \fbox{\includegraphics[width=0.22\linewidth]{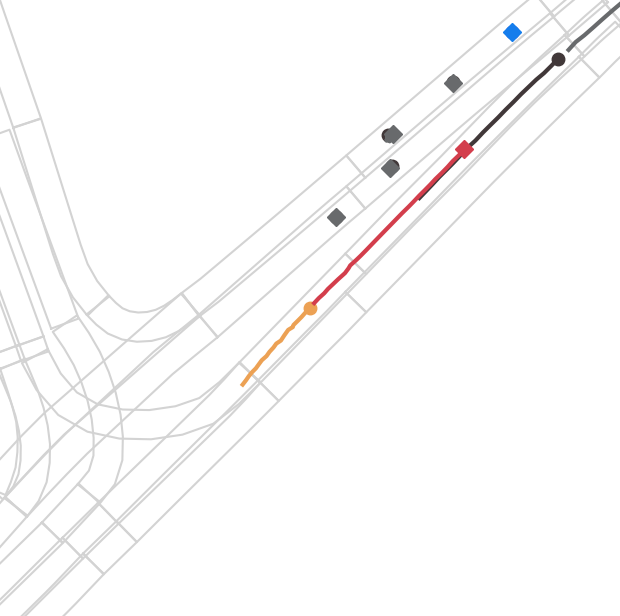}} & 
    \fbox{\includegraphics[width=0.22\linewidth]{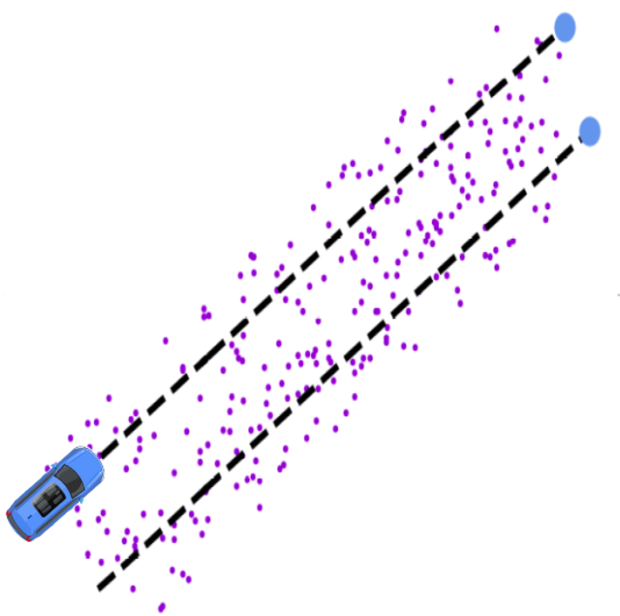}} &
    \fbox{\includegraphics[width=0.22\linewidth]{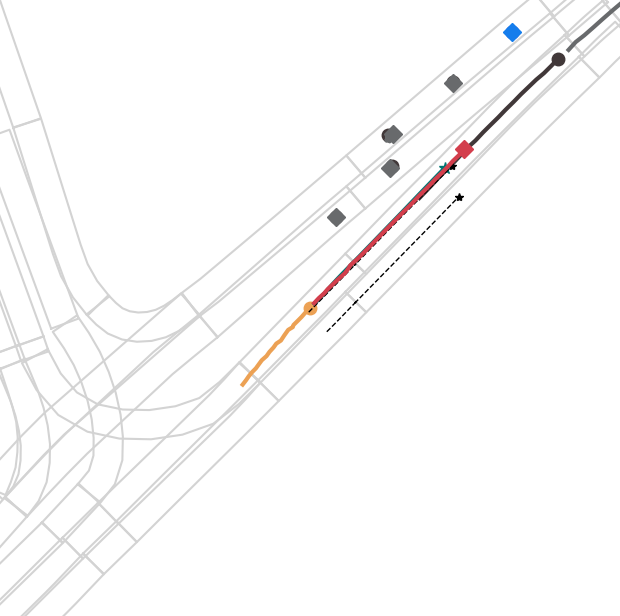}} & 
    \fbox{\includegraphics[width=0.22\linewidth]{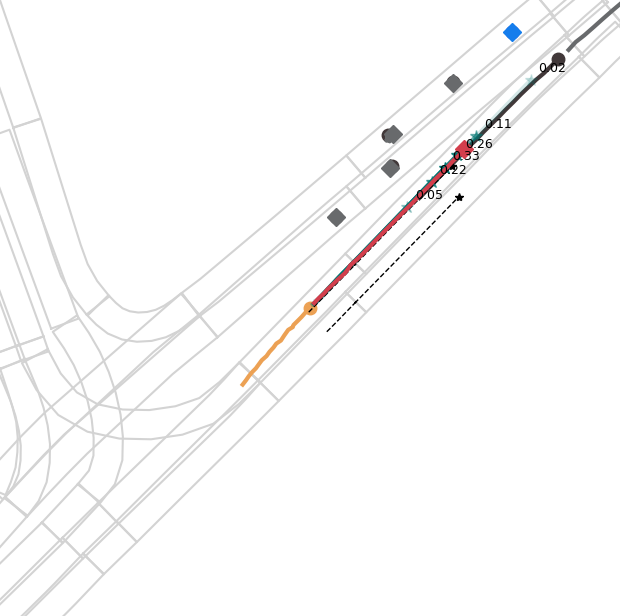}}
    
    \tabularnewline
    
    General view & Plausible HDMap & Unimodal (k=1) prediction & Multimodal (k=6) predictions \tabularnewline
    \end{tabular}
    \caption{Qualitative Results on challenging scenarios using our best model. 
    We represent: our vehicle (\textbf{\textcolor{blue}{ego}}), the \textbf{\textcolor{YellowOrange}{target agent}}, and \textbf{\textcolor{dgray}{other agents}}. We can also see the \textbf{\textcolor{red}{ground-truth}} trajectory of the target agent, our \textbf{\textcolor{ForestGreen}{multimodal predictions}} (with the corresponding confidences) and \textbf{plausible centerlines}. Circles represent last observations and diamonds last future positions.
    As we can see the plausible HDMap serves as a good guidance to our model, which can predict reasonable trajectories in presence of multiple agents and challenging scenarios. We show, from left to right, a general view of the traffic scenario (including social and map information), our calculated plausible HDMap, unimodal prediction (best mode in terms of confidence) and multimodal prediction (\textit{k} = 6), including confidences (the higher, the most probable)}
    \label{fig:results}
\end{figure*}

Finally, we include our low-level (static) features as a static deep physical context which is common to all iterations over the different centerlines and an additional vector distance to the corresponding centerline, achieving our best results without additional regularization terms (hard-mining and Hinge / WTA losses). In both cases (social and map baselines), we obtain regression metrics (minADE and minFDE with both $k$ = 1 and 6) up-to-pair with other SOTA models with a noticeable lower number of parameters, specially in the ablation study for map-based MP models, demonstrating how focusing on the most important map-features drastically decreases the network complexity obtaining similar results in terms of accuracy. This representation not only gathers information about the feasible area around the agent, but also represents potential goal points \cite{dendorfer2020goalgan} (i.e. potential destinations or end-of-trajectory points for the agents). Moreover, this information is \textit{"cheap"} and \textit{interpretable}, therefore, we do not need further exhaustive annotations from the HD Map in comparison with other methods like HOME, which gets as input a 45-channel encoded map \cite{gilles2021home}. We prove this qualitatively in Fig. \ref{fig:results}, where we can see how the estimated centerlines represent a good guidance for the model. 

\paragraph{Comparison with the State-of-the-Art}

The Argoverse Benchmark~\cite{chang2019argoverse} has over 290 submitted methods, however, the top approaches achieve, in our opinion, essentially the same performance. In order to do a fair comparison, we analyze the \textit{state-of-the-art} performance in this benchmark, we show the results in Table~\ref{table:results_test}. Given the standard deviations (in meters) of the most important regression metrics (minADE and minFDE, both in the unimodal and multimodal case), we conclude that there are no significant performance differences for the top-25 models. In fact, Argoverse 2 \cite{wilson2023argoverse}. explicitly mentions that there is a \textit{"goldilocks zone"} of task difficulty in the Argoverse 1 test set, since it has begun to plateau. \\

\begin{table*}[h]
  \centering
  \caption{\textbf{Results on the Argoverse 1.0 Benchmark~\cite{chang2019argoverse}}. We borrow some numbers from~\cite{chang2019argoverse, gilles2021home, gilles2021gohome}. We specify the map info for each model: Raster, GNN or polyline, as stated in Table \ref{table:related_work}. We indicate the error \textcolor{blue}{difference} of our method \emph{w.r.t.} top-25 SOTA methods, in centimeters. Our predictions differ \emph{w.r.t.} top-25 SOTA only \textcolor{blue}{10cm} and \textcolor{blue}{15cm} for the unimodal and multimodal minADE metric respectively, yet our model is much more efficient.}
  \small
  \begin{tabular}{l c c c c c}
    \toprule
    Model & Map info & \multicolumn{2}{c}{K=1} & \multicolumn{2}{c}{K=6}\\
    & & minADE $\downarrow$ & minFDE $\downarrow$ & minADE $\downarrow$ & minFDE $\downarrow$ \\
    \midrule
    Constant Velocity~\cite{chang2019argoverse} & - & 3.53 & 7.89 &  &  \\ 
    Argoverse Baseline (NN)~\cite{chang2019argoverse} & - & 3.45 & 7.88 & 1.71 & 3.29 \\
    Argoverse Baseline (LSTM)~\cite{chang2019argoverse} & Polyline & 2.96 & 6.81 & 2.34 & 5.44  \\
    Argoverse Baseline (NN)~\cite{chang2019argoverse} & Polyline & 3.45 & 7.88 & 1.71 & 3.29  \\
    \midrule
    TPNet-map-mm~\cite{fang2020tpnet} & Raster & 2.23 & 4.70 & 1.61 & 3.70 \\
    Challenge Winner: uulm-mrm (2nd)~\cite{chang2019argoverse} & Polyline & 1.90 & 4.19 & 0.94 & 1.55 \\
    Challenge Winner: Jean (1st)~\cite{mercat2020multiattentmotion, chang2019argoverse} & Polyline & 1.74 & 4.24 & 0.98 & 1.42 \\
    TNT~\cite{zhao2020tnt} & GNN & 1.77 & 3.91 & 0.94 & 1.54 \\
    mmTransformer~\cite{liu2021multimodal} & Polyline & 1.77 & 4.00 & 0.84 &  1.33 \\
    HOME~\cite{gilles2021home} & Raster & 1.72 & 3.73 & 0.92 & 1.36 \\
    LaneConv~\cite{deo2018convolutionalmotion} & Raster & 1.71 & 3.78 & 0.87 & 1.36 \\
    UberATG~\cite{liang2020learninggraph} & GNN & 1.70 & 3.77 & 0.87 & 1.36 \\
    LaneRCNN~\cite{zeng2021lanercnn} & GNN & 1.70 & 3.70 & 0.90 & 1.45 \\
    GOHOME~\cite{gilles2021gohome} & GNN & 1.69 & 3.65 & 0.94 & 1.45 \\
    \textbf{State-of-the-art (top-10)}~\cite{gilles2021gohome, liu2021multimodal, varadarajan2021multipath++, ye2021tpcn} &  & \textbf{1.57}$\pm$0.06 &  \textbf{3.44}$\pm$0.15 & \textbf{0.79}$\pm$0.02 & \textbf{1.17}$\pm$0.04  \\
    \textbf{State-of-the-art (top-25)}~\cite{gilles2021gohome, liu2021multimodal, varadarajan2021multipath++, ye2021tpcn} &  & \textbf{1.63}$\pm$0.08 & \textbf{3.59}$\pm$0.20 & \textbf{0.81}$\pm$0.03 & \textbf{1.22}$\pm$0.06  \\
    \midrule
    Ours (Social baseline, including HardM and losses) & - & 2.57 & 4.36 & 1.26 & 2.67 \\
    \rowcolor{gray} Ours (Map baseline, including HardM and losses) & Polyline & 
    1.73~{\scriptsize{\textcolor{blue}{(10cm)}}} & 3.89~{\scriptsize{\textcolor{blue}{(30cm)}}} & 0.96~{\scriptsize{\textcolor{blue}{(15cm)}}} & 1.63~{\scriptsize{\textcolor{blue}{(41cm)}}} \\
    \bottomrule
  \end{tabular}
  \label{table:results_test}
\end{table*}

On the other hand, in terms of \textbf{efficiency discussion}, to the best of our knowledge, very few methods reports efficiency-related information \cite{gilles2021gohome, gilles2021home, liu2021multimodal, gao2020vectornet}. Furthermore, comparing runtimes is difficult, as only a few competitive methods provide code and models. The Argoverse Benchmark~\cite{chang2019argoverse} provides insightful metrics about the model's performance, mainly related with the predictions error. However, there are no metrics about efficiency (i.e. model complexity in terms of parameters or FLOPs). In the AD context, we consider these metrics as important as the error evaluation because, in order to design a reliable AD stack, we must produce reliable predictions on time, meaning the inference time (related to model's complexity and inputs) is crucial. SOTA methods already provide predictions with an error lesser than 1 meter in the multi-modal case. In our opinion, an accident will rarely happen because some obstacle predictions are offset by one or half a meter, this uncertainty in prediction can be acceptable in the design of AV, but rather because lack of coverage or delayed response time. Despite its high accuracy and fast inference time, LaneGCN \cite{liang2020learninggraph} makes use of multiple GNN layers that can lead to issues with over-smoothing for map-encoders \cite{li2018deeper}. Moreover, as mentioned in \cite{gao2020vectornet}, CNN-based models for processing the HD map information are able to capture social and map interactions, but most of them are computationally too expensive.  

We show the \textbf{efficiency comparison} with other relevant methods in Table \ref{table:effcomp}. 
The results for the other methods are consulted from \cite{gilles2021home} \cite{gilles2021gohome} \cite{gao2020vectornet} \cite{he2022multi}. 
In order to calculate the FLOPs, we follow the common practice \cite{gao2020vectornet} \cite{gu2021densetntwaymo} \cite{gilles2021gohome} of fixing the number of lanes \emph{i.e.,} the number of centerlines is limited to 3. 
%
Gao \emph{etal.} \cite{gao2020vectornet} compares its GNN backbone with CNNs of different kernel sizes and map resolution to compute deep map features (decoder operations and parameters are excluded, min), demonstrating how CNN based methods noticeably increase the amount of parameters and operations per second. We do not require CNNs to extract features from the HD map since we use our map-based feature extractor to obtain the feasible area (see Sec.~\ref{subsec:map_baseline}). 
Moreover, these features are interpretable in comparison with CNNs high-dimensional outputs. Note that, in both variants (social and map baselines), the self-attention module is used with a dynamic number of input agents, this typically implies a quadratic growth in complexity with the number of agents in the scene~\cite{vaswani2017attention}, yet, this only applies to the MHSA layers.

\paragraph{Final Discussion} Even though our methods do not obtain the best regression metrics, we achieve up-to-pair results (Table~\ref{table:effcomp}) against other SOTA approaches whilst our number of FLOPs is several orders of magnitude smaller than other approaches \cite{gu2021densetntwaymo} \cite{liang2020learninggraph}, obtaining a good trade-off (specially the map baseline) between model complexity and accuracy (minADE, k=6), making it suitable for real-time operation in the field of AD. In our case, considering the top-25 regression metrics we achieve near SOTA results (just 15 cm, which represents 18.5 \%, worse in terms of minADE k=6 regarding our final approach) while achieving an impressive reduction of parameters and FLOPs. As observed in Table \ref{table:effcomp}, if we compare our final model, which includes social information, agents interaction and preliminary road information, and the methods with the closest minADE k=6 [m] (LaneGCN \cite{liang2020learninggraph}, HOME \cite{gilles2021home} and GOHOME \cite{gilles2021gohome}), we obtain a reduction of 96 \%, 99 \% and 48 \% respectively in terms of FLOPs. It can be observed how including preliminary road information assuming non-holonomic \cite{triggs1993motion} and anisotropic \cite{ross1989planning} constraints respectively (that is, we mostly focus on the front driveable area) instead of processing the whole map, as well as computing social interactions via graph convolutional networks, boost our model for further integration edge-computing devices with a minimum accuracy loss acceptable for real-world Autonomous Driving applications.

\begin{table}[!ht]
    \centering
    \caption{\textbf{Efficiency comparison among SOTA methods}. We show the number of parameters for each model, FLOPs and minADE (k=6) in the test set. Works from \cite{gao2020vectornet} focus on unimodal predictions (k=1).
    }
    \small
    \resizebox{\linewidth}{!}{
    \begin{tabular}{lccccc}
        \toprule
         Model & \# Params.~(M) & FLOPs (G)~$\downarrow$ & minADE (k=6)~(m) $\downarrow$ \\
         \midrule
         CtsConv~\cite{walters2020trajectory} & 1.08 & 0.34 & 1.85 &  \\
         R18-k3-c1-r400~\cite{gao2020vectornet} & 0.25 & 10.56 & 2.16 \\
         VectorNet~\cite{gao2020vectornet} & \textbf{0.072} & 0.41 & 1.66 \\
         DenseTNT \cite{gu2021densetntwaymo} & 1.1 & 0.763 & 0.88 \\
         LaneGCN \cite{liang2020learninggraph} & 3.7 & 1.071 & 0.87 \\
         mmTransformer \cite{liu2021multimodal} & 2.607 & 0.177 & 0.84 \\
         MF-Transformer \cite{he2022multi} & 2.469 & 0.408 & \textbf{0.82} \\
         HOME~\cite{gilles2021home} & 5.1 & 4.80 & 0.92 \\
         GOHOME~\cite{gilles2021gohome} & 0.40 & 0.09 & 0.94 \\
         \midrule
         Ours (social) & 0.105 & \textbf{0.007} & 1.26  \\
         Ours (social+map) & 0.459 & 0.047 & 0.96  \\
         \bottomrule
    \end{tabular}}
    \label{table:effcomp}
\end{table}


\section{Conclusions and Future Works}
\label{sec:conclusion}

Motion Prediction (MP) requires efficient solutions. Using HD maps and past trajectories leads to the best performance, for this reason, public models use complex processing to leverage both sources of information.

In this work, we propose several baseline models for the well-known Argoverse 1 Motion Predicition Benchmark using state-of-the-art techniques and a novel map model as an alternative to black-box CNN-based map processing. The proposed methods use map-based feature extraction to improve performance by feeding interpretable prior-knowledge, computed by means of classic kinematic techniques, into the model, demonstrating how these prior features help to achieve competitive results in both accuracy and efficiency with a low computational cost compared to other SOTA proposals.

In future work, we will implement of our model in other MP datasets, such as NuScenes, Waymo or Argoverse 2 (which is currently the largest MP dataset in the AD field, beating remaining datasets in terms of unique roadways, average tracks per scenario, total time, diversity and evaluated object categories), to study the generalization and domain adaptation of our algorithms, focusing on Multi-Agent prediction, instead of only Single Agent Prediction. Furthermore, Continual Learning (also referred as Life-Long or Incremental Learning) will be studied by training with smaller training datasets and progressively incorporate enhanced interpretable map features such as dynamic map elements (e.g. traffic light state, goal points) and static map elements, like lane mark types and boundaries or intersections as preliminary physical context information, conducting a fusion-cycle to perform message-passing among the different agents and finally integrating the future centerlines, as done in the present work, to refine the preliminarily decoded multimodal trajectories.


%
We hope that our ideas can serve as a reference for designing efficient MP methods. 

\section*{Acknowledgment}
{\small
This work has been funded in part from the Spanish MICINN/FEDER through the
Artificial Intelligence based modular Architecture Implementation and Validation for Autonomous
Driving (AIVATAR) project (PID2021-126623OB-I00), Electric Automated Vehicle for Aging Drivers (AVAD) project (PDC2022-133470-I00) and from the RoboCity2030-DIH-CM project (P2018/NMT- 4331), funded by Programas de actividades I+D (CAM), cofunded by EU Structural Funds and Scholarship for Introduction to Research activity by University of Alcalá.

Marcos Conde is with the University of Würzburg, supported by the Humboldt Foundation.
}




{\small
\bibliographystyle{IEEEtran}
\bibliography{bibliography}
}




\begin{IEEEbiography}
    [{\includegraphics[width=0.9in,height=1.1in,clip,keepaspectratio]{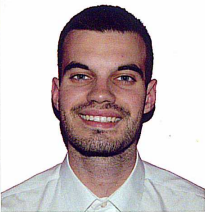}}]{Carlos Gómez-Huélamo}
received his BSc degree (2013-2017) in Industrial Electronics and Automation Engineering and his MSc (2017-2019) in Industrial Engineering, focused on Robotics and Perception, from the University of Alcalá. As a PhD candidate (2019 - ) in Robotics and Artificial Intelligence in the RobeSafe research group (Department of Electronics, University of Alcalá), his PhD thesis is focused on "Predictive Techniques for Scene Understanding by using Deep Learning in Autonomous Driving".
\end{IEEEbiography}

\vspace{-50 pt}

\begin{IEEEbiography}
    [{\includegraphics[trim={5cm 3cm 3cm 0}, clip, width=1.0in]{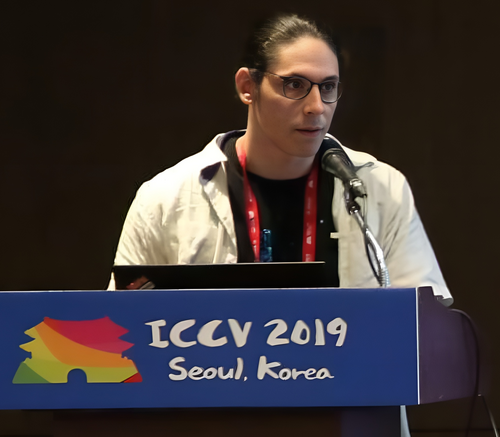}}]{Marcos V. Conde}
received the M.Sc. degree in Computer Vision from the Autonomous University of Barcelona (UAB). He is a PhD researcher at the Computer Vision Lab, University of Würzburg, Germany. He is advised by Prof. Radu Timofte. He is co-organizer of the NTIRE workshop at CVPR. He also serves as Reviewer for top conferences and journals such as CVPR, ICCV, AAAI, NeurIPS, and IEEE TIP/TPAMI. His research interests include artificial intelligence (AI), neural networks, computer vision, image processing, and inverse problems.

\end{IEEEbiography}

\vspace{-50 pt}

\begin{IEEEbiography}
    [{\includegraphics[width=0.9in,height=1.1in,clip,keepaspectratio]{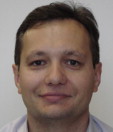}}]{Rafael Barea}
received a Ph.D. degree in Telecommunications from the University of Alcalá in 2001, a MSc degree in Telecommunications from the Technical University of Madrid, Spain, in 1997, and a BSc degree in Telecommunications Engineering from the University of Alcalá in 1994. He is currently Full-Professor (since 2017) in the Department of Electronics at the University of Alcalá. His research interests include Bioengineering, Autonomous Vehicles, Computer Vision and System Control. He is the author of numerous refereed publications in international journals, book chapters and conference proceedings.
\end{IEEEbiography}

\vspace{-50 pt}

\begin{IEEEbiography}
    [{\includegraphics[width=0.9in,height=1.1in,clip,keepaspectratio]{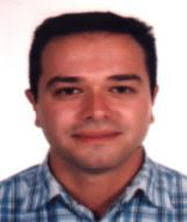}}]{Manuel Ocaña}
is a Associate Professor at the Department of Electronics of the University of Alcalá. He received the Tech. Eng. degree in Electrical Engineering in 2000 from the Technical University of Madrid (UPM), Masters’ degree in Electrical Engineering in 2002 (with Best Student Award) and the Ph.D. degree in Electrical Engineering (with PhD UAH Award) in 2005 from the University of Alcalá (UAH), Alcalá de Henares, Madrid, Spain. His research interests include localization and navigation in Robotics, eSafety and Intelligent Transportation Systems. He is the author of more than 110 refereed papers in journals and international conferences, and corresponding author of 3 patents. 
\end{IEEEbiography}

\vspace{-50 pt}

\begin{IEEEbiography}
    [{\includegraphics[width=0.9in,height=1.1in,clip,keepaspectratio]{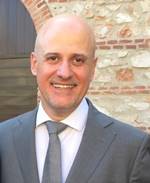}}]{Luis M. Bergasa}
 received the PhD degree in Electrical Engineering in 1999 from the University of Alcalá (UAH), Madrid, Spain. He is Full Professor since 2011 and Director of Digital Transformation since 2022 in this university. From 2000 he had different research and teaching positions at the UAH. He was Head of the Department of Electronics (2004-2010), coordinator of the Doctorate program in Electronics (2005-2010), Director of Knowledge Transfer at the UAH (2014-2018), and Director of the Committee for the Strategic Plan of the UAH (2019-2022). He is author of more than 280 refereed papers in journals and international conferences. His research activity has been awarded/recognized with 28 prizes/recognitions related to Robotics and Automotive fields from 2004 to nowadays. He ranks 65th among Spanish researchers in Computer Science (2022). He was recognized as one of the most productive authors in Intelligent Transportation Systems (ITS) (1996-2014). He was a Distinguished Lecturer of the IEEE Vehicular Technology Society (2019 - 2021). He received the Institutional Lead Award 2019 from the IEEE ITS Society for the longstanding work of his research group. He is Associate Editor of the IEEE Transactions on ITS from 2014 and he has served on Program/Organizing Committees in more than 20 conferences. His research interests include driver behaviors and scene understanding using Computer Vision and Deep Learning for Autonomous Driving.
\end{IEEEbiography}

\end{document}